\documentclass[12pt]{iopart}

\usepackage{graphicx} 
\usepackage{cite}
\usepackage[version=4]{mhchem}
\usepackage{graphicx}
\usepackage{framed} 
\usepackage{amssymb,amsfonts}
\usepackage{cleveref}
\usepackage{caption}
\usepackage{subcaption}
\usepackage[table,xcdraw]{xcolor}
\usepackage{moresize}
\usepackage[normalem]{ulem}
\usepackage[hyphens]{url}
\usepackage{breakurl}

\usepackage{algorithm}
\usepackage{algorithmic} 

\graphicspath{{figures/}}

\begin{document}

\title{Combining SNNs with Filtering for Efficient Neural Decoding in Implantable Brain-Machine Interfaces}
\author{Zhou Biyan,
        Pao-Sheng Vincent Sun,
	and~Arindam~Basu$^{*}$}
\address{City University of Hong Kong, Hong Kong}
\address{B. Zhou and P.S.V. Sun have contributed equally}
\address{$^{*}$ Author to whom any correspondence should be addressed}
\ead{arinbasu@cityu.edu.hk}
\date{June 2024}
\maketitle

\begin{abstract}
While it is important to make implantable brain-machine interfaces (iBMI) wireless to increase patient comfort and safety, the trend of increased channel count in recent neural probes poses a challenge due to the concomitant increase in the data rate. Extracting information from raw data at the source by using edge computing is a promising solution to this problem, with integrated intention decoders providing the best compression ratio. Recent benchmarking efforts have shown recurrent neural networks to be the best solution. Spiking Neural Networks (SNN) emerge as a promising solution for resource efficient neural decoding while Long Short Term Memory (LSTM) networks achieve the best accuracy. In this work, we show that combining traditional signal processing techniques, namely signal filtering, with SNNs improve their decoding performance significantly for regression tasks, closing the gap with LSTMs, at little added cost. Results with different filters are shown with Bessel filters providing best performance. Two block-bidirectional Bessel filters have been used--one for low latency and another for high accuracy. Adding the high accuracy variant of the Bessel filters to the output of ANN, SNN and variants provided statistically significant benefits with maximum gains of $\approx 5\%$ and $8\%$ in $R^2$ for two SNN topologies (SNN\_Streaming and SNN\_3D). Our work presents state of the art results for this dataset and paves the way for decoder-integrated-implants of the future. 
\end{abstract}

\begin{flushleft}\label{AbbreviationsList}
\footnotesize{
    \textbf{List of Abbreviations- } 
    
    \begin{table}[h]
        \begin{tabular}{ll}
            iBMI    & Implantable Brain Machine Interface\\
            NHP    & Non-human Primate\\
            SNN    & Spiking Neural Network\\
            ANN     & Artificial Neural Network \\
            LSTM    & Long Short Term Memory
        \end{tabular}
    \end{table}
}
\end{flushleft}

\section{Introduction}
\begin{figure*}[t]
    \begin{framed}
        \centering
        \includegraphics[width=0.98\textwidth]{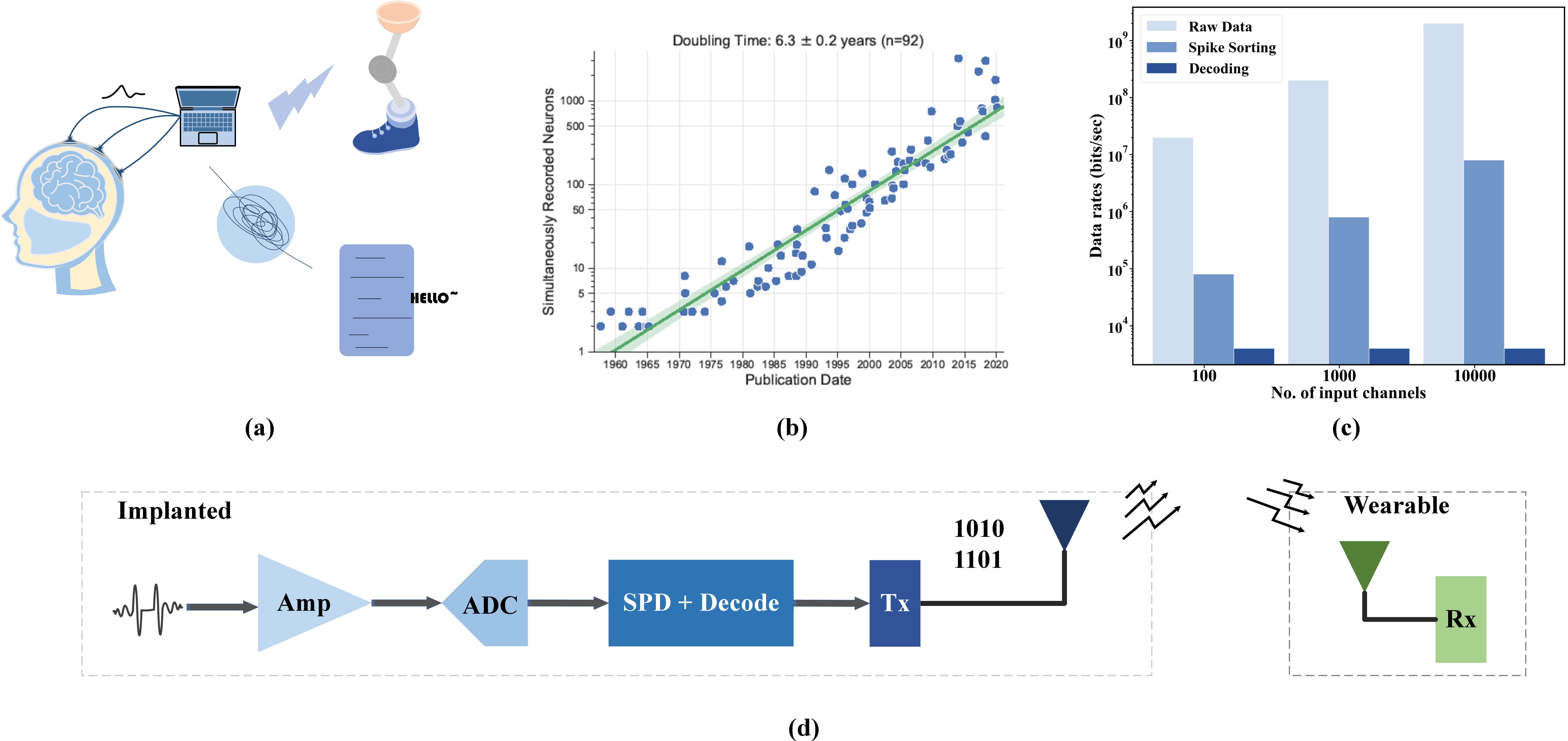}
        \caption{(a) Conceptual figure of an iBMI that reads user intent and controls an effector. (b) Trend of exponential increase in the number of simultaneously recorded neurons\cite{stevenson_2013}. (c) Transmission data rates for the different cases (Raw data, spike sorting, decoding). While spike sorting provides some compression, decoding on the implant can provide the best option, especially as the number of electrodes increases beyond 1000. (d) Integrating computing in the implant can reduce the wireless datarate enabling scalability of iBMI systems.}
    \label{fig:fig1}
    \end{framed}
\end{figure*}

Implantable Brain-Machine Interfaces (iBMI) (\Cref{fig:fig1}(a)) are a promising class of assistive technology that enables the reading of a person's intent to drive an actuator\cite{milin_nature}. It holds promise to enable paralyzed patients to perform activities of daily living with partial or total autonomy\cite{milin_nature}. While the first applications were in motor prostheses to control a cursor on a computer screen\cite{bmi_cursor}, or wheelchairs\cite{camilo_plos}, or robotic arms\cite{bmi_arm}, recent studies have shown remarkable results for speech decoding\cite{speech_decode_nature1,speech_decode_nature2}, handwritten text generation\cite{Brain2Text} and therapies for other mental disorders\cite{MentalBMI}. 

The majority of clinical iBMI systems have a wired connection from the implant to the outside world \cite{milin_nature} that restrict's the user's mobility. Many studies have found user independence to be a top priority for patients\cite{collinger2013functional}. In addition, the wired connection also entails a risk of infection leading to an increasing interest in wireless neural interfaces \cite{Neuralink,milin_nature,nurmikko_wireless}. Another trend in the field has been the constant increase in the number of electrodes\cite{electrodeScaling} to increase the number of simultaneously recorded neurons (\Cref{fig:fig1}(b)) which can increase the precision of decoding the intent of the user and enable dexterous control. The recently developed Neuropixels technology has increased the number of recorded neurons to $\approx 1000$. This can be problematic for wireless implants due to the conflicting requirements of high data rate and low power consumption \cite{murmann_nature}.
Hence, there are efforts to compress the neural data on the sensor by extracting information from it by edge computing (\Cref{fig:fig1}(c,d)). 

Different degrees of computing can be embedded in the implant, from spike detection, classification, to decoding\cite{basu2017big}. When the occurrence of spikes is only of interest, spike detection methods can be used to only transmit the occurrence of spikes and can provide compression rations between 100-1000x (depending on firing rate and signal to noise ratio) compared to the conventional data rate \cite{basu2017big, chae2009128}. However, this method removes all waveform information that could enable the detection of source neurons necessary in neuroscientific experiments. Spike sorting is another compression method in which in addition to the spike, an identifier of the firing neurons is also sent, resulting in slightly reduced compression rates\cite{basu2017big, gibson2011spike}. Ideally, decoding on implant can provide the maximum compression \cite{shaikh2019towards} with the added benefit of patient privacy since the data does not need to leave the implant—only motor commands are sent out. \Cref{fig:fig1}(c) compares the transmission data rate of three methods \cite{chen2015128, shaikh2019towards} assuming 100/1000/10000 channels, neural firing rate of 100 Hz, two classes and decoder output rate of 250 Hz. As the number of channels increases, decoding offers the best compression, as its output data rate is fixed (albeit at the cost of increased decoder complexity). Traditional decoders have used methods from statistical signal processing such as Kalman filters and their variants\cite{dataset, an2022power}. With the rapid growth of Artificial Neural Networks (ANN) and variants for many different applications due to their natural ability to model nonlinear functions and availability of special hardware for training, it is natural to explore the usage of such techniques for motor decoding and several such works have recently been published \cite{shaikh2019towards, willsey2022real, shaeri202433, zhong202433,cindy_tbcas}.

To fit on the implant, the decoder has to be extremely energy and area efficient, along with being accurate. Some specialized decoder integrated circuit using ANNs have been developed to achieve this purpose\cite{chen2015128,shaeri202433,cindy_tbcas}. Brain-inspired SNN are supposed to be even more energy-efficient due to their event-driven nature \cite{liao2022energy, boi2016bidirectional, snn_review}. They are also expected to be better at modeling signals with temporal dynamics due to their inherent “stateful” neurons with memory. However, detailed comparisons between SNN and ANN variants with controlled datasets and benchmarking procedures have been lacking. A recent effort\cite{Yik2023} has put together a benchmarking suite to address this gap and one chosen task is that of motor decoding. We use the same dataset for benchmarking and show additional results for more control cases.

Neurobench showed that streaming SNNs provide a good tradeoff in terms of accuracy vs computes while other methods could have similar memory footprint. Another recent work\cite{hueber2024benchmarking} using the same benchmarking suite showed recurrent networks providing the best results with LSTMs outperforming SNNs in terms of accuracy. We make the following novel contributions in this paper:
\begin{itemize}
\item We improve SNNs for regression by filtering the output using a  Bessel filter from signal processing and close the gap in accuracy with LSTMs.
\item We demonstrate that filtering the outputs of both ANN and SNN decoders with block Bidirectional Bessel filters improves decoding accuracy.
\item We demonstrate state-of-the-art decoding accuracy on this benchmark using LSTMs and filtered SNNs which occupy the higher and lower ends of the pareto optimal curve respectively.
\item We show the effect of increasing training data that shows which models have potential for improvements in future.
\item We show improved accuracy when trained with better curated data; this points to an automated method of trial selection for training neural networks for neural decoding.
\end{itemize}

The rest of the paper is organized as follows. The following section discusses some of the related works while \Cref{sec:Methodology} describes the dataset, models and pre-processing used in this work. \Cref{sec:Results} presents the results comparing different models in terms of their performance-cost tradeoff using pareto curves. This is followed by a \Cref{sec:Discussion} that discusses the main results and provides additional control experiments. Finally, we summarize our findings and conclude in the last section. 

\section{Related Works and Contribution}
\label{sec:Related_works} 

The current work on designing decoders for motor prostheses can be divided into two broad categories--those using traditional signal processing methods and more recent ones based on machine learning.

\subsection{Traditional Signal Processing Decoders}
An early decoder used in BMI system is the linear decoder, such as population vector (PV) algorithm \cite{georgopoulos1986neuronal}. Optimal linear estimators (OLE), generalized from PV algorithm, has comparable performance in closed-loop BMI systems, Whereas Bayesian algorithms perform better \cite{koyama2010comparison}. Inspired by estimation and communication theory, Wiener filter improved linear decoders by combining neuron history activation \cite{kim2008neural}.

Kalman filter has an outstanding ability to cope with dynamic and uncertain environments and is suited in real-time applications. That makes Kalman filter one of the most widely used decoding algorithms in iBMI systems. However, the conventional Kalman filter is only optimal for linear variables and Gaussian noise \cite{dataset}. Many variants of Kalman filter have been proposed to be applied to different applications or environments, such as decoding for cursor movement \cite{dataset}, predicting the movement for clinical devices \cite{murmann_nature}, controlling the robotic arms \cite{hochberg2012reach}, speech decoding \cite{speech_decode_nature1}.

\subsection{Machine Learning Decoders: Algorithms}
Machine learning is widely used in various applications due to its powerful ability to process complex data. An SVM decoder could be trained to analyze rhythmic movements of Quadriplegia patients \cite{sharma2016using}, or motor control of paralyzed limbs \cite{friedenbergneuroprosthetic}. Recently, ANNs have attracted much attention among machine learning algorithms and have made great progress in BMI decoding. ELM-based intelligent intracortical BMI $($ $i^{2}$ BMI $)$ achieves an outstanding performance compared to traditional signal processing decoders \cite{shaikh2019towards}. A multi-layer ANN is trained to decode the finger movement running in a real-time BMI system, which outperforms a Kalman filter \cite{willsey2022real}. 

Recurrent neural networks (RNN) were introduced since they are more skilled at capturing the relations between two variables using a hidden state with memory. For instance, there have been studies on decoding speech \cite{speech_decode_nature2} and on brain representation for handwriting \cite{Brain2Text}. 
Long-term decoding achieved higher performance by using LSTM and Wiener filter \cite{ZZ_FRM_SPD}. To decode speech for a paralyzed person, a natural-language model and Viterbi decoder are used \cite{speechBMI}.

Neuromorphic algorithms have emerged as an energy-efficient decoder and an effective tool for data compression \cite{Schuman2022}. SNN is a brain-inspired neural network popular in neuromorphic applications due to its low energy. It can achieve nearly the same accuracy as ANN but with less than $10\%$ memory access and computation of ANN \cite{Liao2022}. Similarly, it was found that the SNN decoders could use far fewer computes compared to ANN, but with a performance penalty in accuracy, for the motor prediction of primates in the Neurobench benchmark suite\cite{Yik2023}. Another recent work \cite{hueber2024benchmarking} showed that SNNs offer an advantage of low-latency which is essential for closed-loop neuromodulation.

In this work, we show that the combination of traditional signal processing filters with SNNs results in one of the best decoders for iBMI systems.

\subsection{Machine Learning Decoders: Hardware}

Specialized hardware implementations are needed to fit machine learning decoders within the strict area and power budgets of implants. One of the earliest works\cite{chen2015128} used a hardware-algorithm co-design approach to exploit statistical variations in analog circuits to make a sub-microwatt decoder based on extreme learning machine, a variant of reservoir computing algorithms. It also used a configurable digital processing second stage to program distinct weights learned for each chip. More recent work\cite{cindy_tbcas} has used general purpose M0 processor and digital matrix acceleration units, where the power efficiency stems from usage of special features called spike band power (SBP) that require much lower sampling rate than conventional spike detection. While these earlier works demonstrated promising decoder hardware, they were not integrated with neural recording amplifiers as a system on chip. This has been a recent focus\cite{shaeri202433} where multiplexed neural recording front-end circuits are integrated with a 31-class decoder for brain to text applications. In this case, the classifier was as simple linear discriminant analysis (LDA) but its performance is enhanced by a preceding feature extraction module that extracts distinct neural codes. While all of these works used traditional ANNs, one example\cite{boi2016bidirectional} used a SNN decoder to perform a closed loop decoding task of moving an object to a desired location using rodents. Based on the decoded outputs in each step, intra-cortical micro stimulation was delivered to close the loop and allow error correction over multiple time steps. This was a multi-component system not optimized for power dissipation. There is significant opportunity to improve SNN hardware and integrate with neural amplifiers to create a system on chip.

\section{Methodology}
\label{sec:Methodology}

List of notations used in this section
\begin{itemize}
    \item $N_{i}$: $i^{th}$ layer's neuron count
    \item $N_{ch}$: Number of Input Probes
    \item $x_{i}$: Computed feature from i-th probe
    \item $T_W$: Bin window duration
    \item $m$: Number of sub-windows in a bin
    \item $St$: Stride size
    \item $s$: Sparsity
    \item $d$: Dropout rate
    \item $T_{GT}$: The fixed time interval for ground truth labels.
    \item $f_{GT}$: Ground truth label frequency ($=1/T_{GT}$)
\end{itemize}

\subsection{Dataset}
\label{subsec:dataset}

The primate reaching dataset chosen for this paper was gathered and released by \cite{o2017nonhuman}, with the six files chosen for Neurobench \cite{Yik2023} being the files of interest. These six files are recordings of two non-human primates (NHP) (Indy and Loco), where each NHP accounts for three files (more details about this choice in \cite{Yik2023}). 

\begin{figure}[!t]
\begin{framed}
    \centering
    \includegraphics[width=0.45\textwidth]{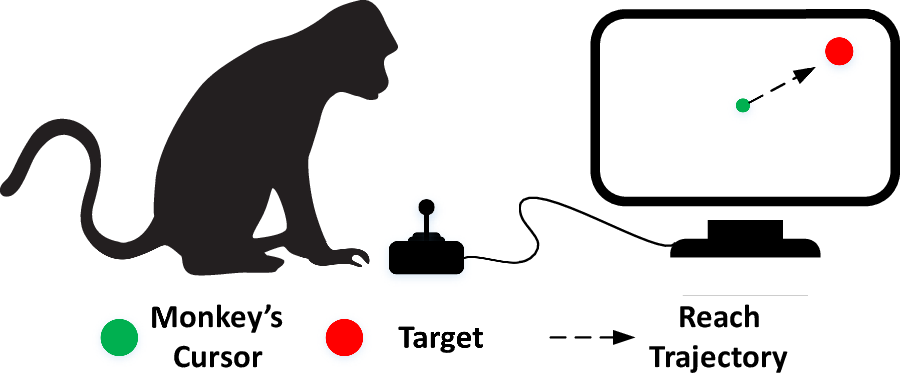}
    \caption{The experiment in the dataset has the NHP controlling the cursor and moving it to the target location. Once the NHP completes the action (referred to as a reach), the target location will move to a new location, and the subject will move the cursor accordingly.}
    \label{fig:pr_dataset}
\end{framed}
\end{figure}

This dataset contains microelectrode array (MEA) recordings of the NHP's brain activity while it is moving a cursor to the target location, as seen in \Cref{fig:pr_dataset}. The finger velocity is sampled at $f_{GT}=250$ Hz resulting in ground truth labels at a fixed interval of $T_{GT}=4$ ms. The target position changes once the monkey successfully moves the cursor to the intended target. We refer to this action as a reach. The dataset contains a continuous stream of the brain's activity from one MEA with $N_{ch}=96$ probes (Indy) or two MEAs with $N_{ch}=192$ probes (Loco). In this work, we ignore sorted spikes since it has been shown that spike detection provides sufficient information for decoding\cite{chen2015128,spd_importance} and is more stable over time. Hence, the number of probes $N_{ch}$ is the input feature dimension $N_0$ for the neural network models (except ANN\_3D) that will be discussed in the following subsection.

\begin{figure}[!t]
\begin{framed}

    \centering
    \begin{subfigure}[b]{0.45\textwidth}
        \centering
        \includegraphics[width=\textwidth]{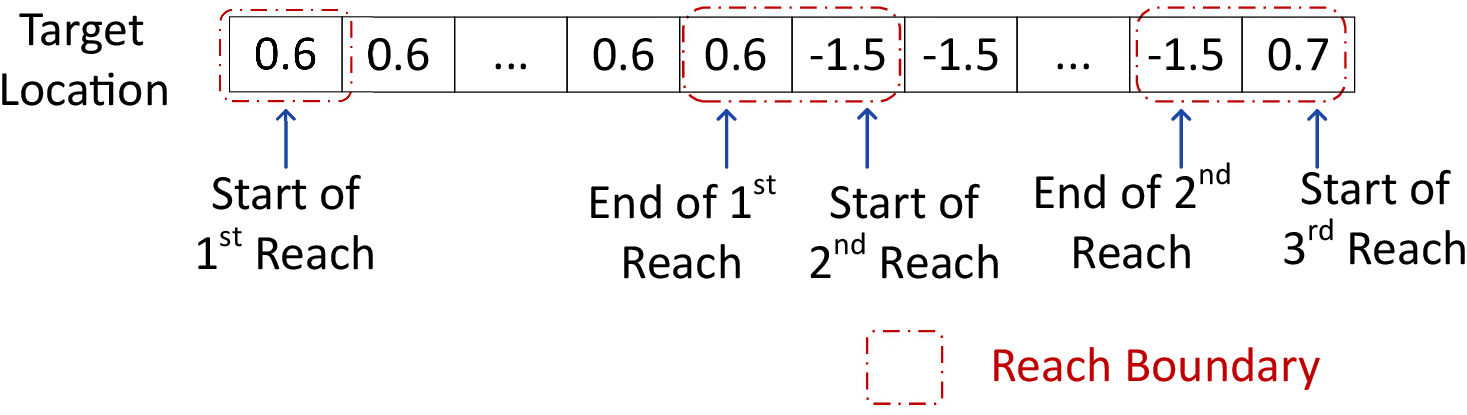}
        \label{fig:summation_method}
        \caption{}
    \end{subfigure}
    \hfill
    \begin{subfigure}[b]{0.45\textwidth}
        \centering
        \includegraphics[width=\textwidth]{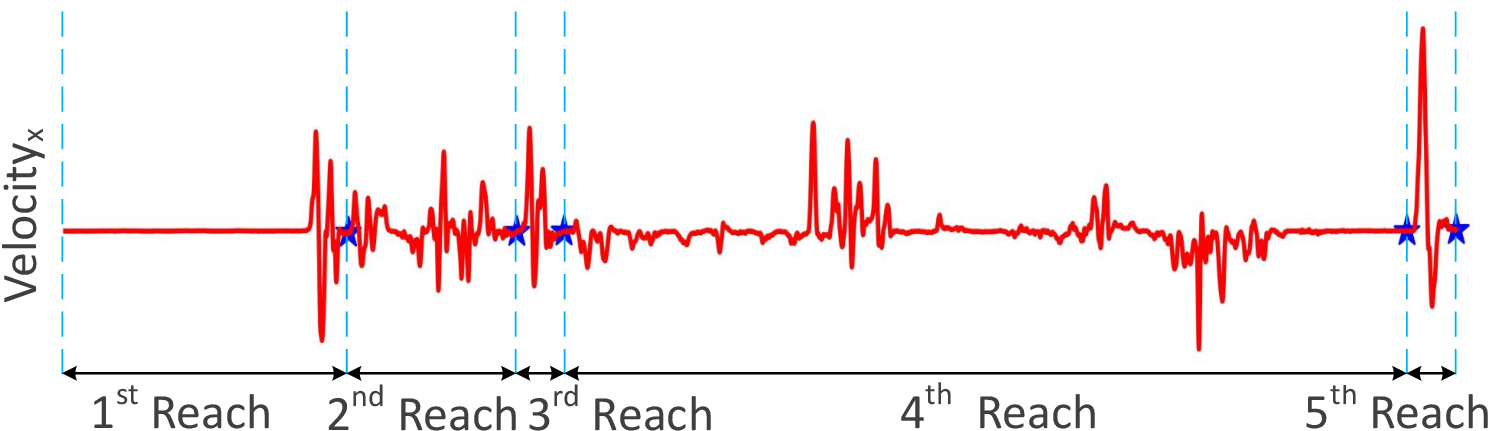}
        \label{fig:step_window_method}
        \caption{}
    \end{subfigure}

    \caption{How each reach is defined in this work: a) The start and end of a reach are marked by the index where there is a change in the target location array, indicating the monkey has moved the cursor to the previous target location. b) A sample segment taken from the file \textit{indy\_20160622\_01}, where we can see five consecutive reaches being segmented.}
    \label{fig:reach_def}
\end{framed}
\end{figure}

Training NN models on time series-based data requires the data to be split apart into separate segments. In analogy with keyword spotting\cite{Yik2023}, each segment of neural data should correspond to separate keywords. By using the target positions in this dataset, we can separate the spike data into segments based on indices in the target position array where there is a change in values, as illustrated in \Cref{fig:reach_def}. Such consecutive indices forms the beginning and end of a reach, and then we can split the time series into training, validation, and test sets based on the number of reaches. The split ratio used in this paper follows that of Neurobench\cite{Yik2023}, which is 50\% for the training set, and 25\% each for validation and test sets. The total number of reaches recorded in each file can be seen in \Cref{tab:num_reaches}.

\begin{table}[h]
\centering
\caption{Number of reaches in each file \cite{zhou2024grand}}
\centering
\footnotesize
\begin{tabular}{@{}llllll}
\br

\centre{1}{\textbf{Filename}} & & \centre{1}{\textbf{Number of Reaches}}   \\
\mr

\centre{1}{\textit{indy\_20160622\_01}} && \centre{1}{970}    \\
\mr

\centre{1}{\textit{indy\_20160630\_01}} && \centre{1}{1023} \\
\mr

\centre{1}{\textit{indy\_20170131\_02}} && \centre{1}{635} \\
\mr

\centre{1}{\textit{loco\_20170210\_02}} && \centre{1}{587} \\
\mr

\centre{1}{\textit{loco\_20170215\_02}} && \centre{1}{409} \\
\mr

\centre{1}{\textit{loco\_20170301\_05}} && \centre{1}{472} \\

\br
\end{tabular}
\label{tab:num_reaches}
\end{table}

\subsection{Network Models}
\label{subsec:networks}

\begin{figure*}[!t]
\begin{framed}
    \centering
    \includegraphics[width=\textwidth]{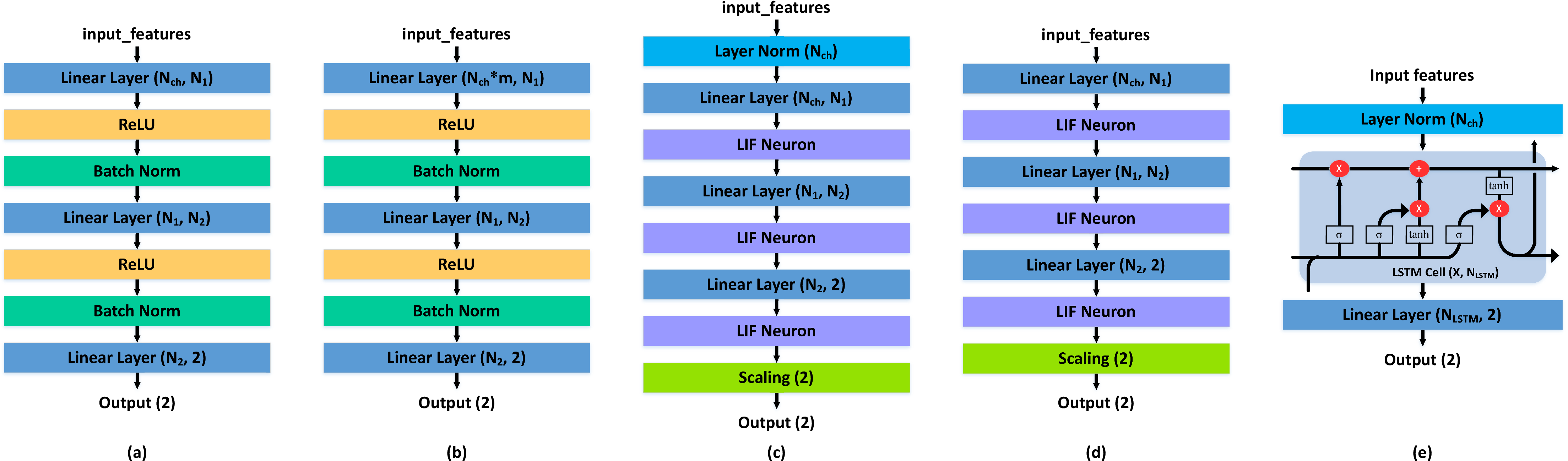}
    \caption{Architecture of models used in this paper are a) ANN b) ANN\_3D c) SNN\_3D d) SNN\_Streaming and e) LSTM.}
    \label{fig:model_arch}
\end{framed}
\end{figure*}

To explore the potential of various neural network models as the neural decoder, five different model architectures with and without memory are tested: ANN, ANN\_3D, SNN\_3D, Streaming SNN and LSTM, which can be seen in \Cref{fig:model_arch}. These five models use NN architectures popular as neural decoders (e.g. ANNs used in \cite{willsey2021real} \cite{glaser2020machine}, SNNs used in \cite{liao2022energy} \cite{taeckens2024spiking} \cite{singanamalla2021spiking} \cite{liao2024spiking} and LSTMs used in \cite{pan2024decoder} \cite{asgher2020enhanced} \cite{premchand2020decoding} \cite{tortora2020deep}) and have memory at the input layer or hidden layer. Every model except for LSTM has two versions of varying complexity (explained in \Cref{subsec:model_search}) where complexity refers to the model size indicating the number of neurons. The larger model is henceforth referred to as the base model while the smaller model is dubbed the \textit{tiny} variant. It was found that networks deeper than 3 layers performed poorly and hence deeper models were excluded from this study.

\begin{enumerate}

    \item \textit{ANN or ANN\_2D}
    
    The \textit{ANN} model has an architecture of $N_{ch}-N_{1}-N_{2}-2$, with rectified linear unit (ReLU) as the activation function for the first two-layers as well as batch normalization to improve upon the accuracy obtained by the model. Note that $N_0=N_{ch}$ indicates one feature extracted from each probe obtained by summing the neural spikes over a fixed duration of $T_W$ as described in \Cref{subsec:Input_processing}. Also, $N_3=2$ corresponds to predicting the $X$ and $Y$ velocities. A dropout layer with a dropout rate of $0.5$ is also added to the first two layers to help regularize the model. In analogy with the naming convention of \textit{ANN\_3D} introduced next, this model can also be referred to as \textit{ANN\_2D} due to the shape of the input weight tensor. 
    
    \item \textit{ANN\_3D or ANN\_flat}
    
    The architecture of the \textit{ANN\_3D} or \textit{ANN\_flat} model is $m\times N_{ch}-N_{1}-N_{2}-2$, i.e. it shares an identical architecture with \textit{ANN}, except at the input layer. This model divides the $T_W$ duration of the input bin window into $m$ sub-windows and creates a $m$-dimensional feature from each probe by summing spikes in each sub-window. This mode of input will be further explained in \Cref{subsec:Input_processing}. The input will then be flattened across the sub-windows, yielding a final input dimension of $N_{ch} \times m$; hence, the number of weights/synapses in the first layer is $m$ times more than \textit{ANN}. It is referred to as \textit{ANN\_flat} in \cite{Yik2023}; we refer to it as \textit{ANN\_3D} here in keeping with the shape of the input weight tensor, which we feel is more intuitive.

    \item \textit{LSTM}
    
    The \textit{LSTM} model contains a single \textit{LSTM} layer of dimension $N_{LSTM}$, followed by a fully-connected layer of dimension $2$. The input of the model shares the same pre-processor as \textit{ANN} (summing spikes in a bin-window of duration $T_W$); however it uses a different $T_{W}$. The input is first normalized with a layer normalization, before passing through the rest of the network.

    \item \textit{SNN\_3D or SNN\_flat}
    
    The \textit{SNN\_3D} aims to achieve high accuracy and shares a similar architecture with \textit{ANN} ($N_{ch}-N_{1}-N_{2}-2$), with the following differences: 1) Instead of using standard activation like ReLU, the \textit{SNN\_3D} model uses the leaky integrate-and-fire (LIF) neuron after every fully-connected layer, 2) the input is first passed through layer normalization, similar to \textit{LSTM} due to the recurrent nature of LIF, 3) at the final layer there is a scaling layer applied to the output LIF neurons and 4) the input spikes are processed using the sub-window method similar to ANN\_3D to capture finer temporal details of the spike rate variations. However, the dimension of the input layer is not $m\times N_{ch}$ like ANN\_3D since in this case, the spikes from the $m$ sub-windows are fed over $m$ time steps to $N_{ch}$ neurons in the first layer using a single weight synapse. The LIF neurons are governed by the following set of equations:
    \begin{align}
        U[t] &= \beta U[t-1] + WX[t] - S_{out}[t-1] \theta \nonumber \\
        \beta &= e^{-\Delta t/\tau} \nonumber \\
        S_{out}[t] &= \begin{cases}
            1 & \text{if } U[t] > U_{thr} \\
            0 & \text{otherwise }
        \end{cases} \nonumber \\
        \theta &= \begin{cases}
            0 & \text{if no reset} \\
            \beta U[t-1] + WX[t] & \text{if reset-to-zero} \\
            U_{sub} & \text{if reset-by-subtraction}
        \end{cases}
        \label{eq:lif_dynamic}
    \end{align}
where $U[t]$ and $X[t]$ are the membrane potential of the LIF neuron and the input at the t-th time step respectively, $W$ is the synaptic weight of the fully-connected layer, $\beta$ is the decay rate, $S_{out}[t]$ is the output spike, $U_{thr}$ is the membrane potential threshold, $U_{sub}$ is the subtracted value if the reset mechanism is reset-by-subtraction and $\theta$ is the reset mechanism. The LIF neurons for all layers shares the same $U_{thr}$ and $\beta$. The first two layer uses the reset-to-zero mechanism while the last layer does not use any reset to allow the final output neurons to accumulate membrane potential to predict the velocity of the primate's movement. For every stride of $4$ ms, the membrane voltages are reset and the integration is restarted with fresh input to produce the next output. Due to the reset of the LIF neurons after every prediction, overlapping bin-windows (for $T_W>St$) cause the \textit{SNN\_3D} to process the same input spikes for multiple predictions.
    
    \item \textit{SNN\_Streaming}

    The \textit{SNN\_Streaming} model also consists of three fully-connected layers ($N_{ch}-N_{1}-N_{2}-2$), with LIF neurons (\Cref{eq:lif_dynamic}) in each layer. Unlike \textit{SNN\_3D}, every LIF layer has its own unique $U_{thr}$ and $\beta$. 
    \textit{SNN\_3D} is designed to achieve the highest accuracy while \textit{SNN\_Streaming} is designed to achieve the best tradeoff between accuracy and resource consumption. Accordingly, the two main differences between \textit{SNN\_3D} and \textit{SNN\_streaming} are the usage of Layer Normalization and the input data processing. SNN\_Streaming avoids using Layer Normalization, which removes data sparsity (by subtracting the mean from the 0 values). The resulting model of SNN\_Streaming has higher sparsity and enables the construction of an energy-efficient model by reducing computations. In terms of input spike processing, $T_W=T_{GT}=St=4$ ms in this model and hence it does not require any additional pre-processing as seen in \Cref{subsec:Input_processing}; hence, it is called a streaming mode since inputs can stream in directly and continuously to this model. Just like \textit{SNN\_3D}, the first two layer uses reset-to-zero while the last layer does not reset its membrane potential.

\end{enumerate}

\subsection{Feature Extraction by Input Spike Processing}
\label{subsec:Input_processing}

\begin{figure*}[!t]
\begin{framed}
    \begin{subfigure}[b]{0.32\textwidth}
        \centering
        \includegraphics[width=\textwidth]{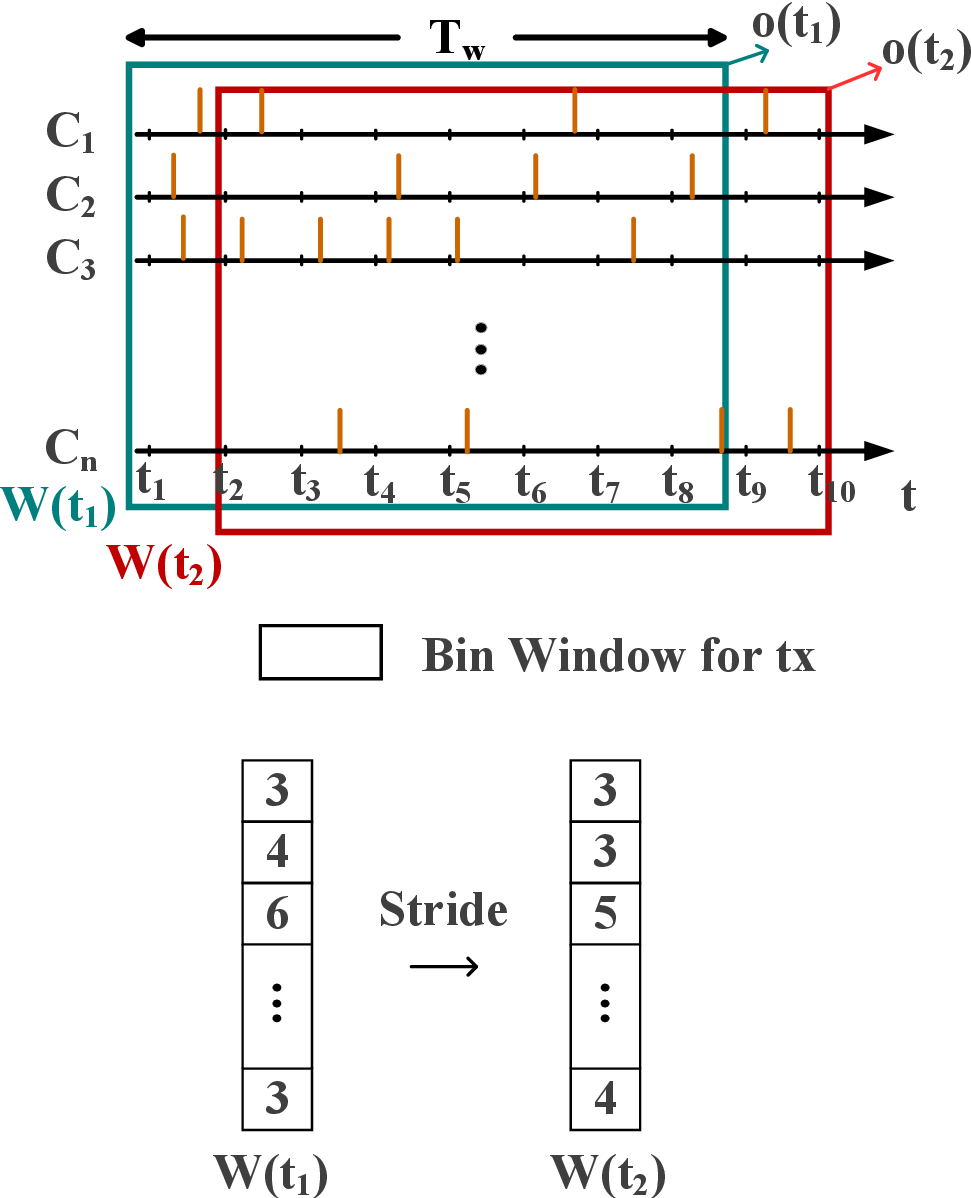}
        \label{fig:summation_method}
        \caption{}
    \end{subfigure}
    \hfill
    \begin{subfigure}[b]{0.32\textwidth}
        \centering
        \includegraphics[width=\textwidth]{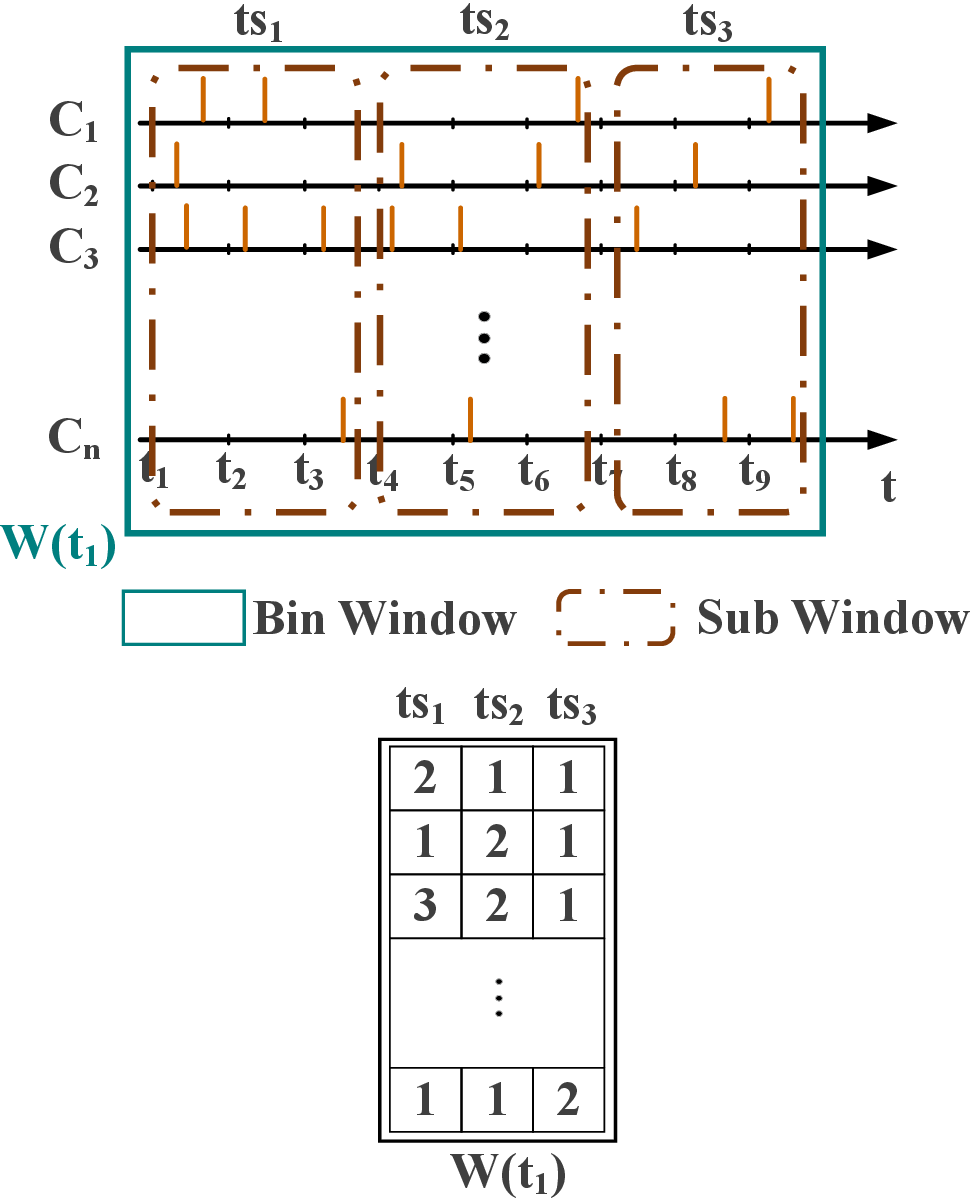}
        \label{fig:step_window_method}
        \caption{}
    \end{subfigure}
    \hfill
    \begin{subfigure}[b]{0.32\textwidth}
        \centering
        \includegraphics[width=\textwidth]{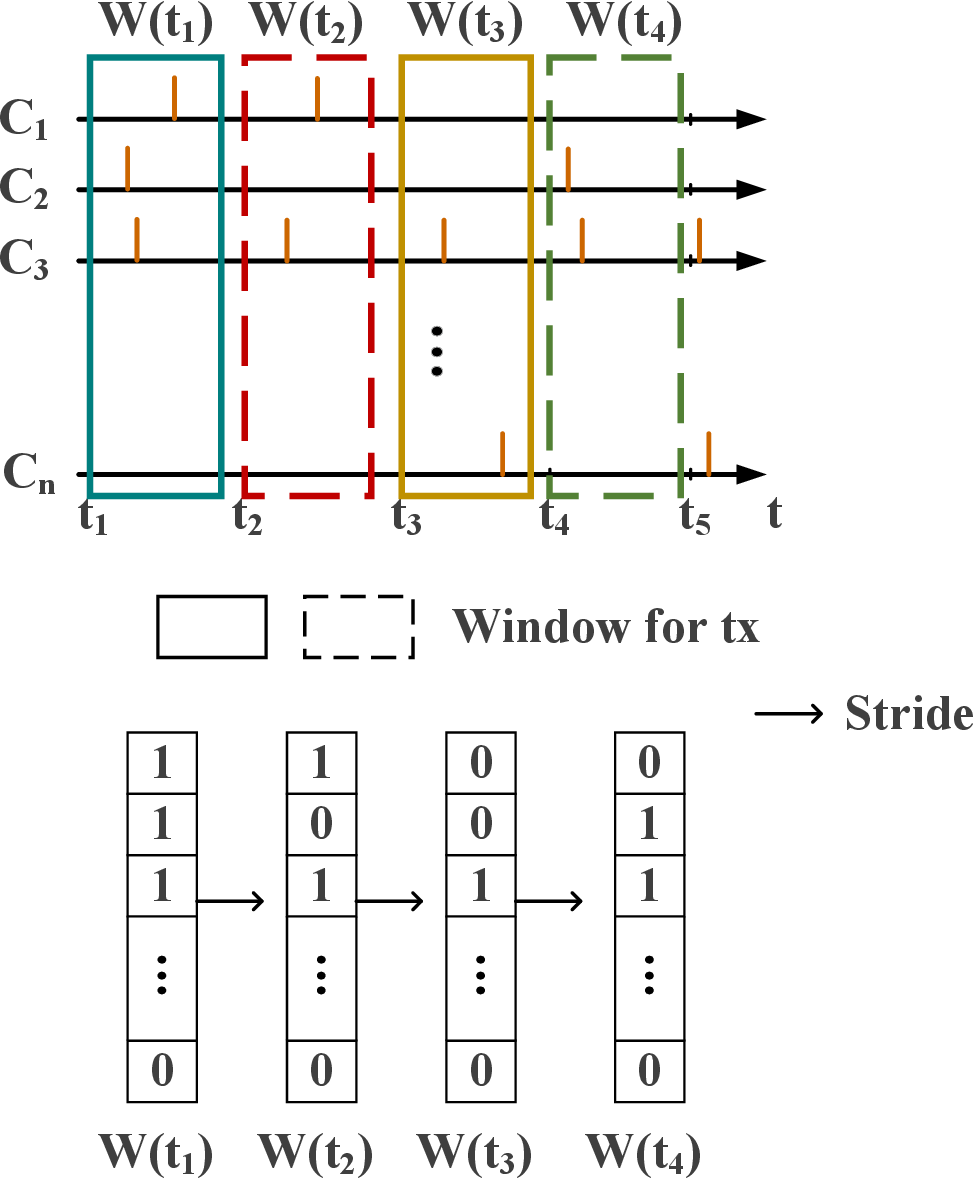}
        \label{fig:streaming_method}
        \caption{}
    \end{subfigure}

    \caption{Input data pre-processing methods for feature extraction presented in this paper: a) Summation mode, where the number of spikes detected within a bin window $T_W$ for each probe is summed to create a feature. b) Sub-window mode, where the bin window is further divided into $m$ sub-windows, and the number of spikes detected within each sub-window is summed. c) Streaming mode, where the input spike is gathered as it is.}
    \label{fig:preprocessing}
    
\end{framed}
\end{figure*}

The spikes generated by the NHP's neurons are sparse in nature. SNNs can intrinsically accept sparse spiking input since they create an accumulation in the membrane potential variable. For ANNs however, the information over a past time period has to be explicitly accumulated in a feature extraction step. Also, from the biological viewpoint, it is generally assumed that short term firing rates (as opposed to mean firing rates over a trial duration \cite{gerstner2014neuronal}) are important for motor control \cite{aggarwal2008asynchronous} \cite{chen2015128}. Hence, we calculate firing rates, $r_i(t_k)$ at the sample time $t_k$ from the spike waveforms $P_i=\sum_{t_{s,i}}\delta (t - t_{s,i})$ on the i-th probe ($1\leq i \leq N_{ch}$) using the following equation:
\begin{equation}
    \label{eq:rate_feature}
    r_i(t_k)=\int_{t_k-T_W}^{t_k}P_i(t)dt
\end{equation}
where $t_{k+1}-t_{k}=T_{GT}$ is the sampling time, $t_{s,i}$ denote neural spike times on the i-th probe and $T_W$ is the bin window duration. Three different pre-processing methods were used in this paper: the summation method, the sub-window method and the streaming method as illustrated in \Cref{fig:preprocessing}. For all of them, the stride size, $st$ is identical to the sampling duration, which is $T_{GT}=4$ ms. They differ in the choice of $T_W$ and how to present the firing information in the bin window to the network as described next.

\begin{enumerate}
    \item Summation Method (used in ANN and LSTM):

This is the simplest case where the firing rate in a bin window with duration of $T_{W}$ is directly used as a feature and input to the NN. We define the input feature vector $\overline{x}(t_k)$ as follows:
    \begin{align}
        \overline{x}(t_k) &= [x_{0}(t_k), x_{1}(t_k), ..., x_{N_{ch}}(t_k)] \nonumber \\
        x_{i}(t_k) &= r_i(t_k)
        \label{eq:summation_method}
    \end{align}
This method is depicted in \Cref{fig:preprocessing}(a). This method is used by \textit{ANN} and \textit{LSTM} models, where \textit{ANN} uses $T_{W}=200$ms while \textit{LSTM} uses $T_{W}=32$ms. Both these bin window sizes were obtained after optimization using the training and validation data. Generally, shorter time windows are preferred to capture fine temporal structure of spike trains\cite{wen2021capturing} and reduce the latency of response \cite{liu2021edge}. However, with short time windows, LSTMs (or other recurrent models) can retain a memory about long-term history through their state variables while ANNs cannot. Hence, the time window for ANN needs to be longer than that of LSTM to retain sufficient information. In terms of hardware realization, while this accumulation of spikes is straightforward for non-overlapping bin windows, cases with overlap would require repeated operations with overlapping data in naive implementations. Efficient implementation of such firing rate calculation with overlapping windows are shown in \cite{chen2015128} using recursion.
    
    \item Sub-Window Method (used in ANN\_3D and SNN\_3D):

    Similar to the summation method, the sub-window method uses information over the latest $T_{W}$ bin window. However, instead of summing all the spikes, it provides firing rate information at an even shorter time-scale (or with finer resolution) of $T_W/m$. Thus, the feature computed from the i-th probe itself becomes a vector $\overline{x_i}(t_k)=[r_i^1(t_k),r_i^2(t_k)...r_i^m(t_k)]$ with $m$ components corresponding to firing rates in each of the $m$ sub-windows (duration of integration in \Cref{eq:rate_feature} is reduced to $T_W/m$). The sub-window method is illustrated in \Cref{fig:preprocessing}(b) and is used by the \textit{ANN\_3D} and \textit{SNN\_3D} models with $T_{W}=200$ms and $m=7$. The feature vector $\overline{x}(t_k)$ for \textit{ANN\_3D} is defined according to \Cref{eq:step_window_method_ANN} as follows:
    \begin{align}
        \overline{x}(t_k) &= [\overline{x_0}(t_k),\overline{x_1}(t_k)..\overline{x_{N{ch}}}(t_k)] 
        \label{eq:step_window_method_ANN}
    \end{align}
    where the dimension of $\overline{x}(t_k)$ is $N_{ch}\times m$. For the \textit{SNN\_3D}, the firing rates in each \emph{sub-window} are given as input feature to the \textit{SNN}, which has $m$ time steps. Thus the input feature vector for the \textit{SNN} in the j-th time step ($1\leq j \leq m$) is given by:
        \begin{align}
        \overline{x_j}(t_k) &= [r_j^1(t_k),r_j^2(t_k)...r_j^{N_{ch}}(t_k)] 
        \label{eq:step_window_method_SNN}
    \end{align}
    where the dimension of $\overline{x_j}(t_k)$ is $N_{ch}$. Note that `j' indexes time steps here and the \textit{SNN} output at $j=m$ is the prediction of motor velocity for sample time $t_k$.
    
    \item Streaming Method (used in SNN\_Streaming):
    The streaming method, as the name suggests, processes the incoming spike data as a continuous stream as seen in \Cref{fig:preprocessing}(c). 
    The streaming method aims to achieve a better tradeoff between accuracy and computations compared to SNN\_3D.
    In this case, $T_W=st=T_{GT}=4$ ms implying no overlap between consecutive windows. This allows for a direct interface between the probes and the model, without the need of adding additional compute cost to our network like the two methods mentioned before. The input feature vector $\overline{x}(t_k)$ is given by the following equation:
    \begin{align}
        \overline{x}(t_k) &= [u(r_0(t_k)),u(r_1(t_k)),...u(r_{N_{ch}}(t_k))] 
        \label{eq:streaming}
    \end{align}
    where $u()$ denotes the Heaviside function. Hence, the resulting \textit{SNN} can replace multiply and accumulate (MAC) operations by selective accumulation (AC) operations.
\end{enumerate}

\subsection{Filters for SNN}
\label{subsec:filter}

\begin{figure}[!t]
\begin{framed}
    \centering
    \includegraphics[width=0.45\textwidth]{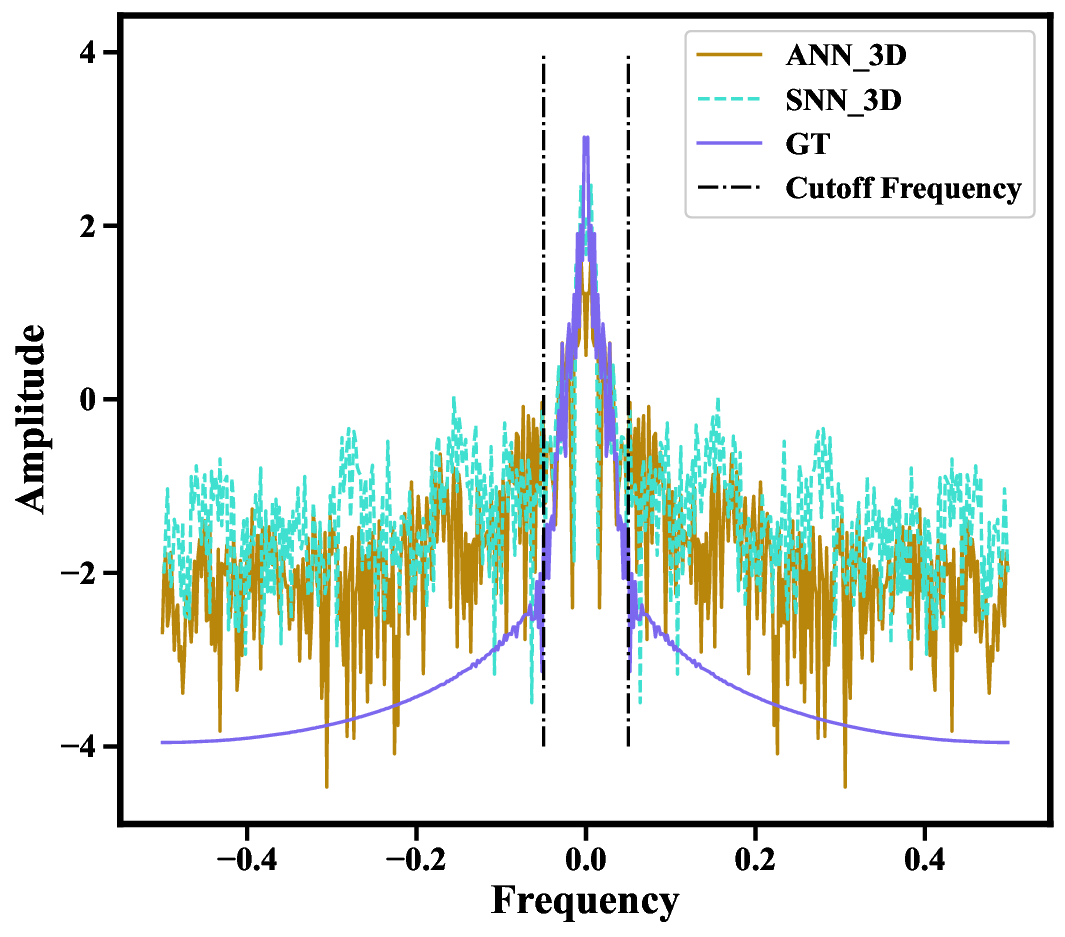}
    \caption{Frequency distribution of ground truth and two model outputs from \cite{Yik2023} for 2 sec of data from the fourth file in \Cref{tab:num_reaches} shows higher frequency content in prediction. This indicates a strategy to estimate the filter cut off frequency.}
    \label{fig:output_frequncy}
\end{framed}
\end{figure}

Most of the NN models (with the exception of LSTM and SNN\_Streaming) introduced in \Cref{subsec:networks} operate on a window or chunk of inputs; providing these windows in any order would result in the same prediction. However, in real life the motor output is a smooth signal with a continuous trajectory. To understand this, we plot in \Cref{fig:output_frequncy} the frequency content of ground truth trajectories of a sample 2-sec waveform and compare it with predicted trajectories of two models from \cite{Yik2023}. It is clear that the predictions have much higher frequency content indicating ground truth trajectories are smoother. In signal processing, this can be rectified by using a filter, which amounts to adding a memory of the past output. 

Three types of filters are compared to further explore the effect of filtering-- Bessel filter, Butterworth filter and Chebyshev filter. Bessel filters provide a linear phase response resulting in constant delay, but it requires higher filter order to achieve same attenuation at high frequency compared to others. Butterworth filter with no ripples in the passband provides a maximum flat response in the passband, which means the filter will introduce minimal variations to the desired signal amplitude. However, the Butterworth filter has a wide transition band, which makes its rolloff gentle, and hence, the Chebyshev filter becomes a third alternative. 

In terms of digital filter implementation, three different techniques were tested in this work. First, we tried forward (Fwd) filtering, which can achieve real-time filtering, but cannot have zero phase shift. The improvements achieved with this method was marginal and we do not consider them in the following parts. On the contrary, bidirectional (Bid) filtering can effectively eliminate phase distortions, but it is generally applicable to offline filtering since the whole waveform is needed before processing begins. To achieve a compromise, block bidirectional filter with a sliding window is applied, such that only a latency penalty of half block size is applicable. We vary the block size between $16-80$, the order of filters between $2-8$ and their cutoff frequencies in the range of $0.05-0.5$ respectively to find the optimum for each model.

\subsection{Metrics}
\label{subsec:metrics}
In order to evaluate the performance of the models comprehensively in terms of cost vs performance, three metrics are used: (1) number of operations, (2) memory footprint, and (3) accuracy. Three types of operations are considered for (1) -- multiply, add and memory read (since the energy for memory access often dominates the energy for computations\cite{horowitz20141}). For most NNs, each synaptic operation comprises a multiply and add (MAC) since the neuron activations and weights are not binary. On the other hand, for SNNs, the synaptic operations only involves accumulations (AC) because of the binary neuron activation. Note that the operations mentioned in this work refer to the operations between synaptic connections as detailed in \Cref{subsec:baseline_result}, while the operations within neurons that determine the membrane potential are excluded. The number of operations is used as proxy for power/energy in this work since the actual energy ratio between these three operations depends on bit-width, process node and memory size; more accurate energy evaluations will be the subject of future work. For (2), memory footprint is evaluated from model size where every parameter is stored using a 32-bit float number. For (3), Coefficient of determination ($R^2$) is a commonly used metrics for regression tasks\cite{dataset,Yik2023,shaikh2019towards}, which is defined by \Cref{eq:r2}: 
\begin{equation}
    R^{2} = 1-\frac{ {\textstyle \sum_{i=1}^{n}\left ( y_{i}-\hat{y_{i} }  \right )^{2}  } }{{\textstyle \sum_{i=1}^{n}\left ( y_{i}-\bar{y} \right )^{2}  }} 
    \label{eq:r2}
\end{equation}
where the label and predictions are showing as $y_{i}$ and $\hat{y_{i}}$ respectively while $ \bar{y}$ is the mean of labels. For motor prediction, separate $R_X^2$ and $R_Y^2$ are computed for predicting X and Y velocities respectively and the final $R^2$ is an average of the two.

Another set of important metrics for NN hardware are throughput and latency. We have not considered them here since the considered NN models are small enough so that the total time taken for evaluating the prediction is dominated by the input data accumulation time shown in \Cref{subsec:Input_processing} and delay due to filtering. However, we do touch upon this point later in \Cref{subsec:filter}.

\subsection{Training \& Testing Details}
\label{subsec:training_details}
All models are trained for $50$ epochs using the SNNTorch framework, with a learning rate of $0.005$, a dropout rate between $0.3-0.5$, and an L2-regularization value between $0.005-0.2$. AdamW is chosen as the optimizer, Mean Squared Error (MSE) loss is determined as the loss function, and a learning scheduler (cosine annealing schedule) is used after every epoch. For ANN, ANN\_3D, and SNN\_3D, data is shuffled with batch size of $512$ in training. For SNN\_3D, the membrane potential resets every batch, while reset occurs at the beginning of each reach for membrane potential in SNN\_Streaming and hidden states in LSTM. The distribution of reach durations show most reaches completed in less than $4$ sec while some reaches being much longer, presumably due to the NHP not attending to the task. Similar to \cite{Yik2023}, reaches that exceed $8$ seconds in length are removed to improve the training performance. Leaky Integrate-and-Fire Neuron is used in SNNs, where the threshold and $\beta$ are learned during training and Arctan is applied as a surrogate function \cite{neftci2019surrogate}. The membrane potential of neurons ceases to reset in the last layer to enable regression. The velocity predicted by SNNs is determined by scaling the membrane potential of neurons with a learnable constant parameter. For validation and testing, data is input to the models in chronological order, and reset mechanisms only occur at the beginning. Filters are employed exclusively during the inference process.

\section{Results}
\label{sec:Results}
To comprehensively examine the capability of different models, we performed multiple experiments and evaluated models using the metrics mentioned in \Cref{subsec:metrics}. All the results except memory access are obtained from the neurobench harness\cite{Yik2023} that does automated evaluation of the models; memory access is estimated based on theoretical equations of weight fetches based on experimentally observed sparsity multiplying the number of weights on a per layer basis. The findings are presented pictorially using two pareto plots, first comparing the accuracy versus operations trade-off and the second comparing accuracy versus memory footprint (e.g. see \Cref{fig:baseline_result} and \Cref{fig:pareto_80_vs_50_split}). 
A tabular summary of all the experiments performed for our base models can be found in \Cref{tab:results_summary}. \Cref{tab:results_summary} also compares the results with other published work using statistical methods such as Steady State Kalman filter (SSKF), Unscented Kalman filter (UKF), recurrent Exponential-Family Harmonium (rEFH) etc.

\begin{table}[t]
\centering
\caption{Baseline Performance and Comparison with Prior Works using \textbf{high accuracy} filter configuration. Best performing filter in a class of SNN models is highlighted in bold font. Incremental accuracy improvement by using $80\%$ data over $50\%$ data is shown in parenthesis.}
\tiny
\begin{tabular}{@{}lllllllll}
\br
&&&&\centre{1}{\textbf{Activation}} &&&& \centre{1}{\textbf{Model Size}} \\

\centre{1}{\textbf{Models}} & \centre{1}{\textbf{Split}} & \centre{1}{\textbf{Filters}} & \centre{1}{\textbf{R$^{2}$}} & \centre{1}{\textbf{Sparsity}} &  & \centre{1}{\textbf{Computes}} &  & \centre{1}{\textbf{$($ kB $)$}} \\

&&&&&\crule{3}\\

&&&&&\centre{1}{\textbf{MACs}} &\centre{1}{\textbf{ACs}} &\centre{1}{\textbf{Memory}}\\

&&&&&&& \centre{1}{\textbf{Access}}\\

\mr

\centre{1}{ANN} & \centre{1}{50\%} & \centre{1}{No Filter \cite{Yik2023}} & \centre{1}{0.5818} & \centre{1}{0.7514} & \centre{1}{4969.76} & \centre{1}{0} & \centre{1}{5,179.46} & \centre{1}{26.5234}\\

&  & \centre{1}{{Block Bid Filtering}} & \centre{1}{{0.6165}} & \centre{1}{{0.7514}} & \centre{1}{{4974.76}} & \centre{1}{{0}} & \centre{1}{{5184.46}} & \centre{1}{{26.5429}}\\

&  & \centre{1}{{Bid Filtering}} & \centre{1}{{0.6168}} & \centre{1}{{0.7514}} & \centre{1}{{4974.76}} & \centre{1}{{0}} & \centre{1}{{5184.46}} & \centre{1}{{26.5429}}\\

&\crule{8}\\

& \centre{1}{80\%} & \centre{1}{No Filter} & \centre{1}{0.6119 \ssmall $($+0.03$)$} & \centre{1}{0.7417} & \centre{1}{5000.25} & \centre{1}{0} & \centre{1}{5,205.65} & \centre{1}{26.5234}\\

&  & \centre{1}{{Block Bid Filtering}} & \centre{1}{{0.6456}} & \centre{1}{{0.7417}} & \centre{1}{{5005.25}} & \centre{1}{{0}} & \centre{1}{{5210.65}} & \centre{1}{{26.5429}}\\

&  & \centre{1}{{Bid Filtering}} & \centre{1}{{0.6461}} & \centre{1}{{0.7417}} & \centre{1}{{5005.25}} & \centre{1}{{0}} & \centre{1}{{5210.65}} & \centre{1}{{26.5429}}\\
 
\mr

\centre{1}{ANN\_3D} & \centre{1}{50\%} & \centre{1}{No Filter \cite{Yik2023}} & \centre{1}{0.6013} & \centre{1}{0.7348} & \centre{1}{11507.07} & \centre{1}{0} & \centre{1}{11,555.31} & \centre{1}{134.5234}\\

&  & \centre{1}{{Block Bid Filtering}} & \centre{1}{{0.6646}} & \centre{1}{{0.7348}} & \centre{1}{{11512.07}} & \centre{1}{{0}} & \centre{1}{{11560.31}} & \centre{1}{{134.5429}}\\

&  & \centre{1}{{Bid Filtering}} & \centre{1}{{0.656}} & \centre{1}{{0.7348}} & \centre{1}{{11512.07}} & \centre{1}{{0}} & \centre{1}{{11560.31}} & \centre{1}{{134.5429}}\\

&\crule{8}\\

& \centre{1}{80\%} & \centre{1}{No Filter} & \centre{1}{0.6523 \ssmall $($+0.05$)$} & \centre{1}{0.7324} & \centre{1}{11676.22} & \centre{1}{0} & \centre{1}{11,644.82} & \centre{1}{134.5234}\\

&  & \centre{1}{{Block Bid Filtering}} & \centre{1}{{0.6859}} & \centre{1}{{0.7324}} & \centre{1}{{11681.22}} & \centre{1}{{0}} & \centre{1}{{11649.82}} & \centre{1}{{134.5429}}\\

&  & \centre{1}{{Bid Filtering}} & \centre{1}{{0.6887}} & \centre{1}{{0.7324}} & \centre{1}{{11681.22}} & \centre{1}{{0}} & \centre{1}{{11649.82}} & \centre{1}{{134.5429}}\\

\mr

\centre{1}{SNN\_3D} & \centre{1}{50\%} & \centre{1}{No Filter \cite{Yik2023}} & \centre{1}{0.6219} & \centre{1}{0} & \centre{1}{32256} & \centre{1}{0} & \centre{1}{39,057.79} & \centre{1}{33.1992}\\

&  & \centre{1}{Block Bid Filtering} & \centre{1}{\textbf{0.6729}} & \centre{1}{0} & \centre{1}{32261} & \centre{1}{0} & \centre{1}{39,062.79} & \centre{1}{33.2187}\\

&  & \centre{1}{Bid Filtering} & \centre{1}{0.6687} & \centre{1}{0} & \centre{1}{32261} & \centre{1}{0} & \centre{1}{39,062.79} & \centre{1}{33.2187}\\

&\crule{8}\\

& \centre{1}{80\%} & \centre{1}{No Filter} & \centre{1}{0.6564 \ssmall $($+0.03$)$} & \centre{1}{0} & \centre{1}{32256} & \centre{1}{0} & \centre{1}{39,701.38} & \centre{1}{33.1992}\\

&  & \centre{1}{Block Bid Filtering} & \centre{1}{\textbf{0.7062 \ssmall  $($+0.03$)$}} & \centre{1}{0} & \centre{1}{32261} & \centre{1}{0} & \centre{1}{39,706.38} & \centre{1}{33.2187}\\

&  & \centre{1}{Bid Filtering} & \centre{1}{0.6909 \ssmall $($+0.02$)$} & \centre{1}{0} & \centre{1}{32261} & \centre{1}{0} & \centre{1}{39,706.38} & \centre{1}{33.2187}\\

\mr

\centre{1}{SNN\_Streaming} & \centre{1}{50\%} & \centre{1}{No Filter} & \centre{1}{0.6112} & \centre{1}{0.7453} & \centre{1}{0} & \centre{1}{971.26} & \centre{1}{1195.28} & \centre{1}{25.32}\\

&  & \centre{1}{Block Bid Filtering} & \centre{1}{0.6449} & \centre{1}{0.7453} & \centre{1}{0} & \centre{1}{976.26} & \centre{1}{1,200.28} & \centre{1}{25.3395}\\

&  & \centre{1}{Bid Filtering} & \centre{1}{\textbf{0.6458}} & \centre{1}{0.7453} & \centre{1}{0} & \centre{1}{976.26} & \centre{1}{1,200.28} & \centre{1}{25.3395}\\

&\crule{8}\\

& \centre{1}{80\%} & \centre{1}{No Filter} & \centre{1}{0.6483 \ssmall $($+0.04$)$} & \centre{1}{0.7795} & \centre{1}{0} & \centre{1}{883.36} & \centre{1}{1,044.23} & \centre{1}{25.32}\\

&  & \centre{1}{Block Bid Filtering} & \centre{1}{\textbf{0.6763 \ssmall $($+0.03$)$}} & \centre{1}{0.7795} & \centre{1}{0} & \centre{1}{888.36} & \centre{1}{1,049.23} & \centre{1}{25.3395}\\

&  & \centre{1}{Bid Filtering} & \centre{1}{0.6761 \ssmall $($+0.03$)$} & \centre{1}{0.7795} & \centre{1}{0} & \centre{1}{888.36} & \centre{1}{1,049.23} & \centre{1}{25.3395}\\

\mr

\centre{1}{LSTM} & \centre{1}{50\%} & \centre{1}{No Filter} & \centre{1}{0.6508} & \centre{1}{0} & \centre{1}{22687.97} & \centre{1}{0} & \centre{1}{22913.27} & \centre{1}{90.95}\\

&  & \centre{1}{{Block Bid Filtering}} & \centre{1}{{0.6711}} & \centre{1}{{0}} & \centre{1}{{22692.97}} & \centre{1}{{0}} & \centre{1}{{22918.27}} & \centre{1}{{90.96}}\\

&  & \centre{1}{{Bid Filtering}} & \centre{1}{{0.6683}} & \centre{1}{{0}} & \centre{1}{{22692.97}} & \centre{1}{{0}} & \centre{1}{{22918.27}} & \centre{1}{{90.96}}\\

&\crule{8}\\

& \centre{1}{80\%} & \centre{1}{No Filter} & \centre{1}{0.6943 \ssmall $($+0.04$)$} & \centre{1}{0} & \centre{1}{22687.97} & \centre{1}{0} & \centre{1}{22912.40} & \centre{1}{90.95}\\

&  & \centre{1}{{Block Bid Filtering}} & \centre{1}{{0.7046}} & \centre{1}{{0}} & \centre{1}{{22692.97}} & \centre{1}{{0}} & \centre{1}{{22918.27}} & \centre{1}{{90.96}}\\

&  & \centre{1}{{Bid Filtering}} & \centre{1}{{0.7051}} & \centre{1}{{0}} & \centre{1}{{22692.97}} & \centre{1}{{0}} & \centre{1}{{22918.27}} & \centre{1}{{90.96}}\\

\mr

\centre{1}{SNN 2D\cite{Yik2023}} & \centre{1}{50\%} & \centre{1}{-} & \centre{1}{0.5805} & \centre{1}{0.9976} & \centre{1}{0} & \centre{1}{413.52} & \centre{1}{1503} & \centre{1}{28.56}\\

\mr

\centre{1}{SNN2\cite{hueber2024benchmarking}} & \centre{1}{50\%} & \centre{1}{-} & \centre{1}{0.6292} & \centre{1}{-} & \centre{1}{-} & \centre{1}{202} & \centre{1}{1815} & \centre{1}{30}\\

\mr

\centre{1}{ELM \cite{shaikh2019towards}} & \centre{1}{50\%} & \centre{1}{-} & \centre{1}{0.5546} & \centre{1}{-} & \centre{1}{217202} & \centre{1}{-} & \centre{1}{217344} & \centre{1}{1140.625}\\

\mr

\centre{1}{rEFH\_dynamic \cite{dataset}} & \centre{1}{{320s$^{*}$}} & \centre{1}{-} & \centre{1}{0.6319} & \centre{1}{-} & \centre{1}{3167} & \centre{1}{-} & \centre{1}{13000} & \centre{1}{229376}\\

&&& \centre{1}{\ssmall $($bin\_width=128$)$}\\

\mr

\centre{1}{SSKF \cite{shaikh2019towards}} & \centre{1}{{50$\%$}} & \centre{1}{-} & \centre{1}{0.1955} & \centre{1}{-} & \centre{1}{426.4} & \centre{1}{-} & \centre{1}{741.6} & \centre{1}{2.48}\\

\mr

\centre{1}{UKF \cite{hueber2024benchmarking}} & \centre{1}{{50$\%$}} & \centre{1}{-} & \centre{1}{0.4510} & \centre{1}{-} & \centre{1}{28799} & \centre{1}{0} & \centre{1}{116000} & \centre{1}{753664}\\
\mr

\centre{1}{LSTM \cite{hueber2024benchmarking}} & \centre{1}{{50$\%$}} & \centre{1}{-} & \centre{1}{0.6109} & \centre{1}{-} & \centre{1}{1154} & \centre{1}{0} & \centre{1}{659000} & \centre{1}{5000}\\

\mr

\centre{1}{{LSTM \cite{glaser2020machine}}} & \centre{1}{{50$\%$}} & \centre{1}{-} & \centre{1}{{0.6746}} & \centre{1}{{0}} & \centre{1}{{872393.04}} & \centre{1}{{0}} & \centre{1}{{872,993.27}} & \centre{1}{{3417.35}}\\

\br
\end{tabular}
\label{tab:results_summary}
\end{table}

\begin{figure}[!t]
\begin{framed}
    \centering
    \includegraphics[width=0.5\textwidth]{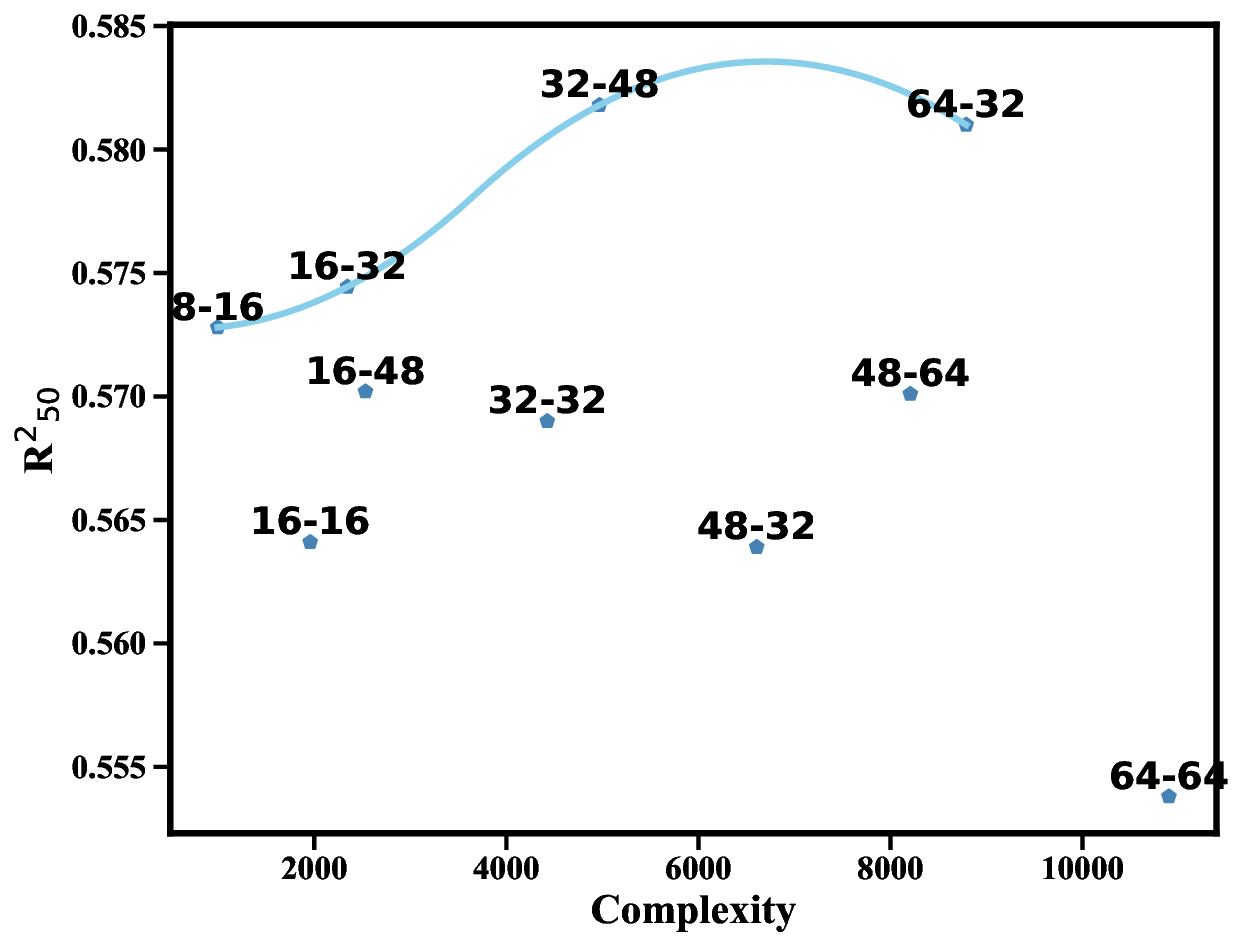}
    \caption{Complexity versus $R^{2}$ for different ANN models tested. $N_1-N_2$ values of $16-32$ and $32-48$ were on the pareto curve and chosen as `tiny' and `baseline' variants. Here, model complexity is characterized by the number of synaptic weights.}
    \label{fig:Complex_vs_r2}
\end{framed}
\end{figure}

\subsection{Model Size Search}
\label{subsec:model_search}
As mentioned earlier, it was found that networks deeper than 3 layers performed poorly and hence deeper models were excluded from this study. The number of neurons in each of the two hidden layers was determined by searching within a certain range ($N_0=N_{ch}$ and $N_3=2$ are fixed). We used the ANN to do this search due to its simplest network structure and resultant fast training. The results obtained by varying $N_1$ and $N_2$ are shown in the \Cref{fig:Complex_vs_r2}. \emph{Here, model complexity is characterized by the number of synaptic weights.} The text shown in the figure represents the different network architectures ($N_1-N_2$ combinations) tested. As expected, the $R^{2}$ initially increases with increasing number of neurons but starts decreasing after the number of neurons reaches a certain value due to overfitting. The best trade-off between $R^{2}$ and complexity is determined by the networks lying on the pareto curve shown in blue in \Cref{fig:Complex_vs_r2}. Therefore, the two models with $N_1-N_2$ values of $32-48$ and $16-32$ were selected as the `base' and `tiny' variants respectively for ANN. Same variant sizes were near optimal for ANN\_3D and SNN\_3D (we do not show these tradeoff curves for brevity), while for SNN\_Streaming, base and tiny variants represented $N_1-N_2$ values of $32-48$ and $16-48$ respectively.

\subsection{K-Fold Cross Validation}
It is important to verify that the result will not vary significantly regardless of how the data is split. Hence, K-fold cross-validation is used to test all six files for three models (ANN, ANN\_3D and SNN\_3D). We divided the data into five parts, randomly selecting four parts as training and the other part was divided into validation and testing. The means and standard deviations of $R^{2}$ for the 5-fold experiment  are shown in \Cref{tab:kfold}. Low variance of the results for all 3 cases implies using one-fold data split for our experiment is reasonable and will give dependable results. As a comparison, the results in \Cref{tab:results_summary} does show that without filter, the decoding accuracy for SNN\_3D is the best and ANN is the worst with ANN\_3D between the two.  Hence, we just use the single data split in \cite{Yik2023} described earlier for the rest of the results.

\begin{table}[h]
\centering
\caption{K-fold Cross Validation}
\centering
\footnotesize
\begin{tabular}{@{}llllll}
\br

\centre{1}{\textbf{Models}} & & \centre{1}{\textbf{$R^{2}$ Mean}} & & \centre{1}{\textbf{$R^{2}$ Standard Deviation}}  \\

\mr
\centre{1}{ANN} && \centre{1}{0.6186}  && \centre{1}{0.0294}    \\
\mr
\centre{1}{ANN\_3D} && \centre{1}{0.6467} && \centre{1}{0.0299}  \\
\mr
\centre{1}{SNN\_3D} && \centre{1}{0.6661}  && \centre{1}{0.0252}  \\
\mr
\centre{1}{SNN\_Streaming} && \centre{1}{0.6144}  && \centre{1}{0.027}  \\
\mr
\centre{1}{LSTM} && \centre{1}{0.6755}  && \centre{1}{0.036}  \\

\br
\end{tabular}
\label{tab:kfold}
\end{table}

\subsection{Filtering: Performance improvement and Optimization}
\label{subsec:filtering}

\begin{figure}[!t]
\begin{framed}
    \centering
    \includegraphics[width=0.65\textwidth]{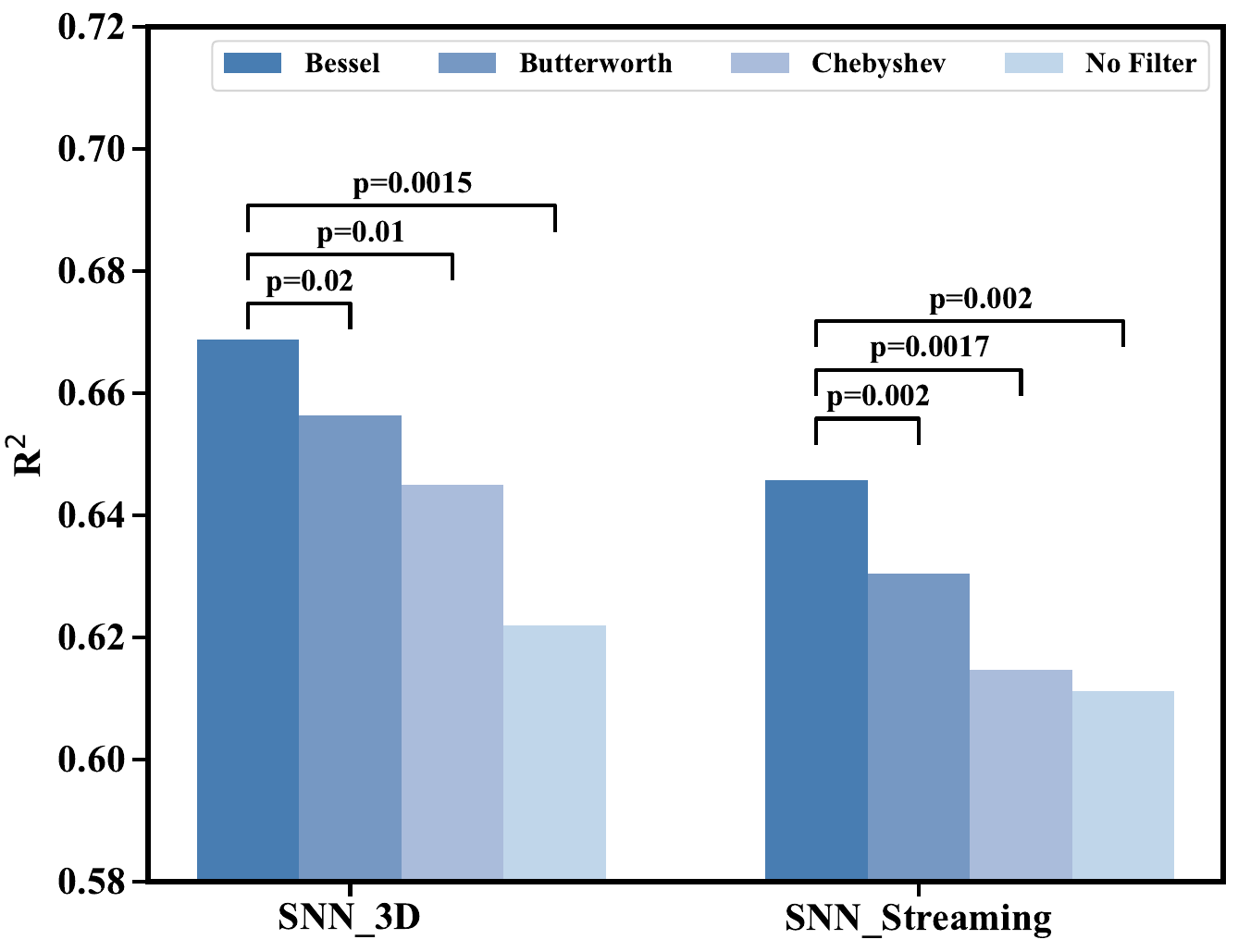}
    \caption{Comparison of different $4$-th order filters using bidirectional technique and cut-off frequency of $0.05$. Bessel filters show the best improvement. Statistical significance tested using paired t-test.}
    \label{fig:filter_compare}
\end{framed}
\end{figure}

\begin{figure}[!t]
\begin{framed}
    \centering
    \includegraphics[width=0.75\textwidth]{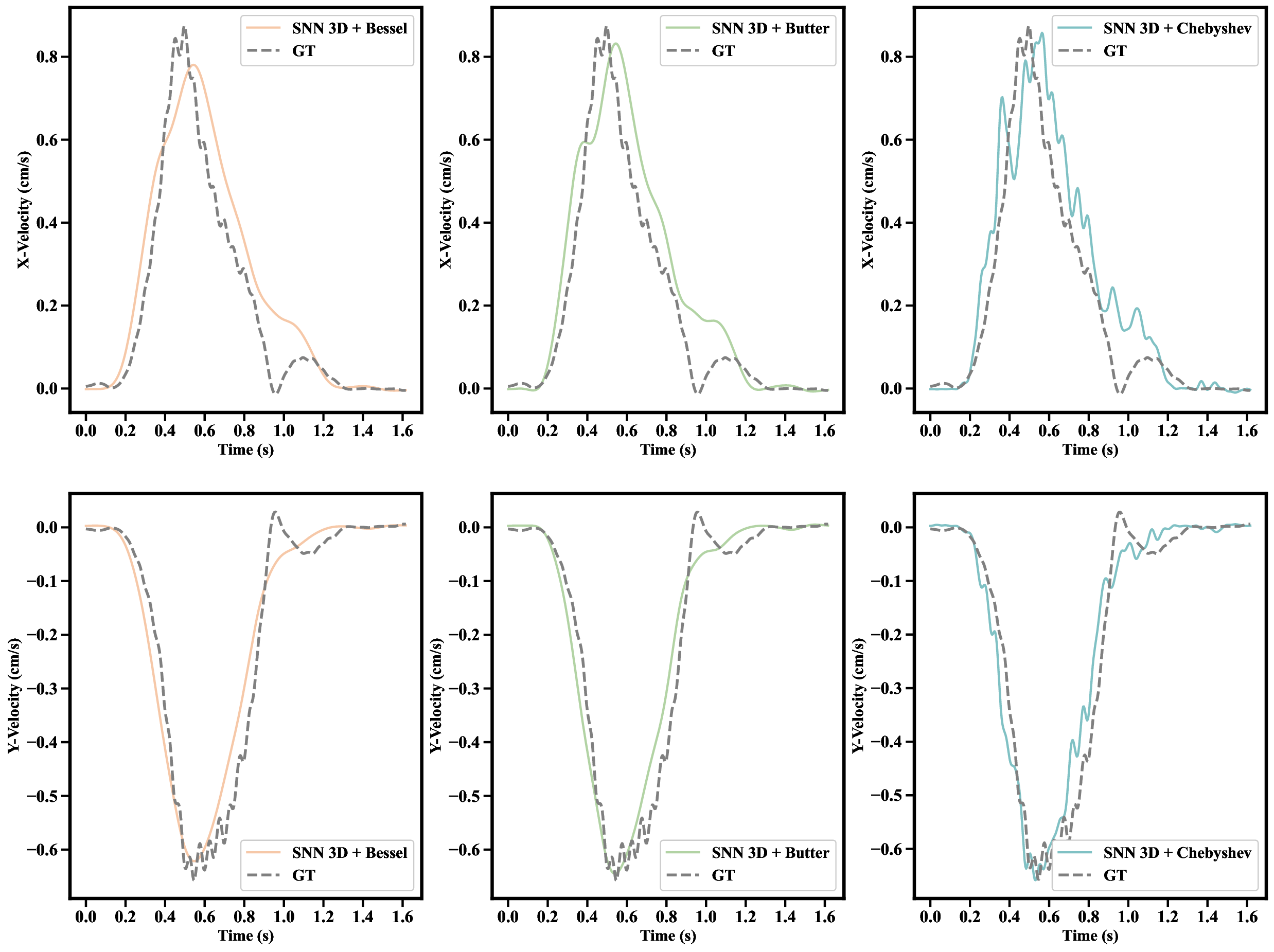}
    \caption{Comparison of three types of filters in terms of velocity and position based on one reach in the file $"indy\_20160622\_01"$. The three columns indicate SNN with Bessel, Butter, and Chebyshev filter respectively. The first row shows the X velocity, the second row shows the Y velocity. 4-th order filter with block size of 32 and cut-off frequency of 0.05 is used.}
    \label{fig:filters_details}
\end{framed}
\end{figure}

\begin{figure}[!t]
\begin{framed}
    \centering
    \begin{subfigure}[b]{0.8\textwidth}
        \centering
        \includegraphics[width=\textwidth]{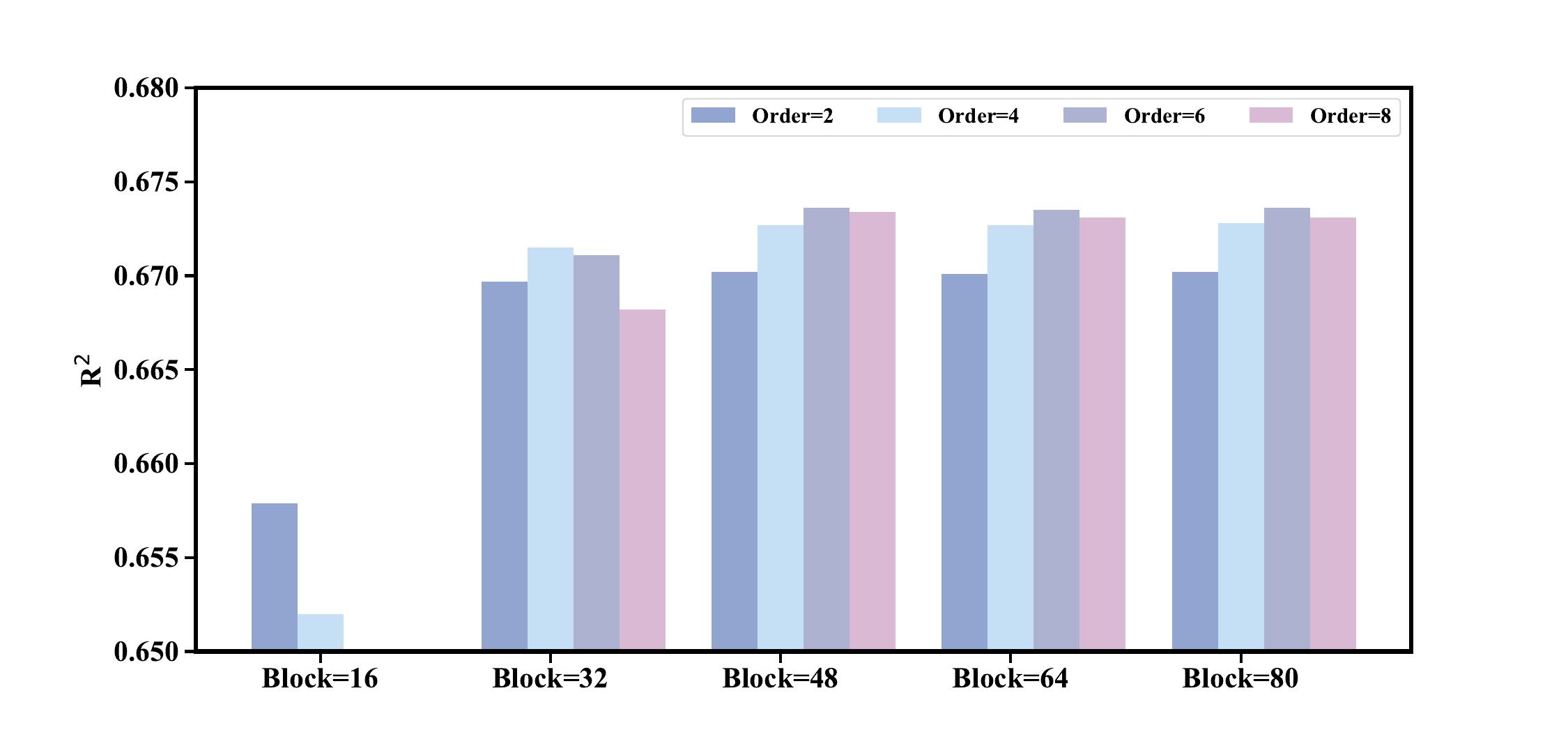}
        \label{fig:3D_order_vs_block_size}
        \caption{}
    \end{subfigure}
    
    \begin{subfigure}[b]{0.8\textwidth}
        \centering
        \includegraphics[width=\textwidth]{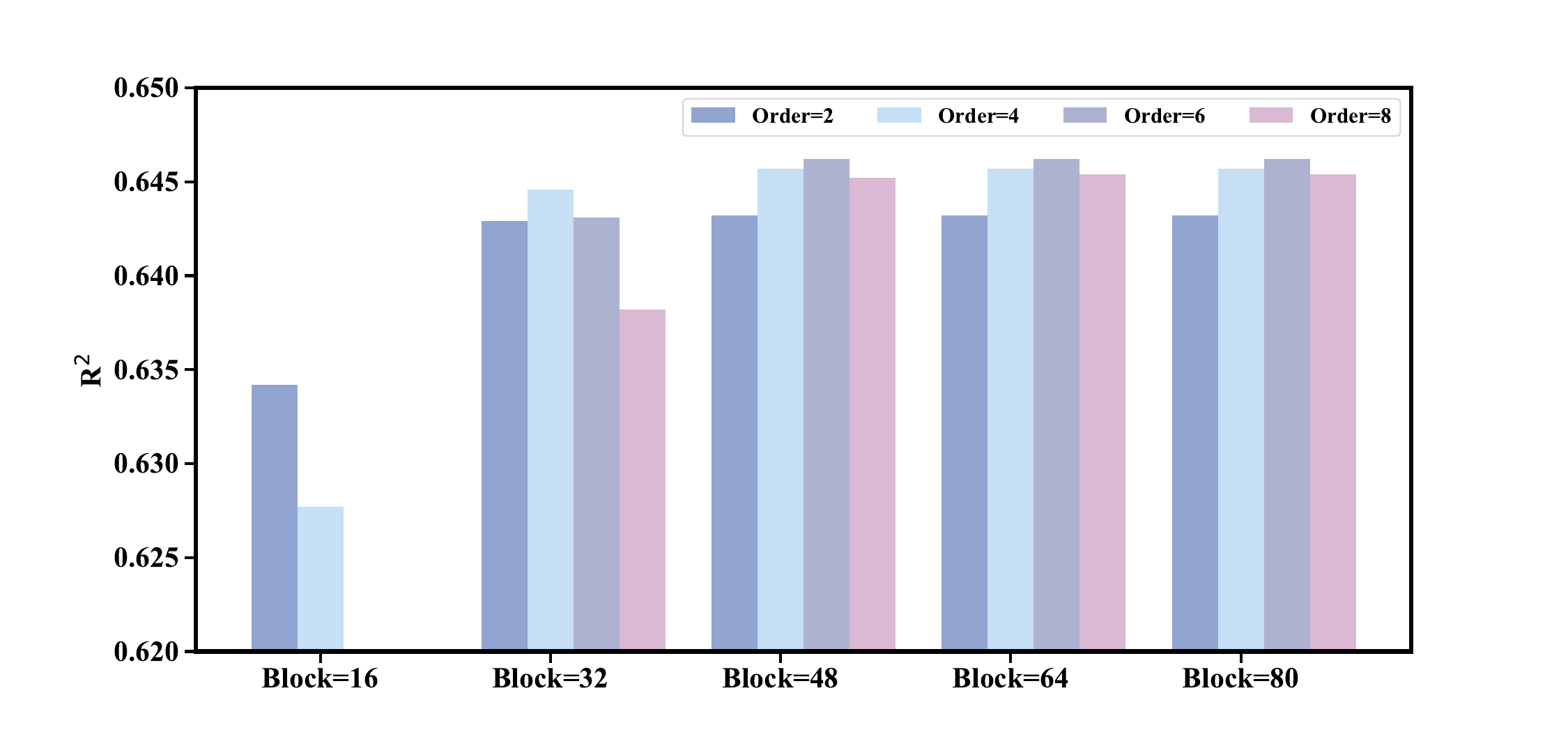}
        \label{fig:streaming_order_vs_block_size}
        \caption{}
    \end{subfigure}

    \caption{Comparison of block size versus $R^{2}$ in different filter order: a) The performance for SNN\_3D using bessel filter with cutoff frequency of 0.05. b) The performance for SNN\_Streaming using bessel filter with cutoff frequency of 0.05.}
    \label{fig:order_vs_block_size}
\end{framed}
\end{figure}

\begin{figure}[!t]
\begin{framed}
    \centering
    \includegraphics[width=0.65\textwidth]{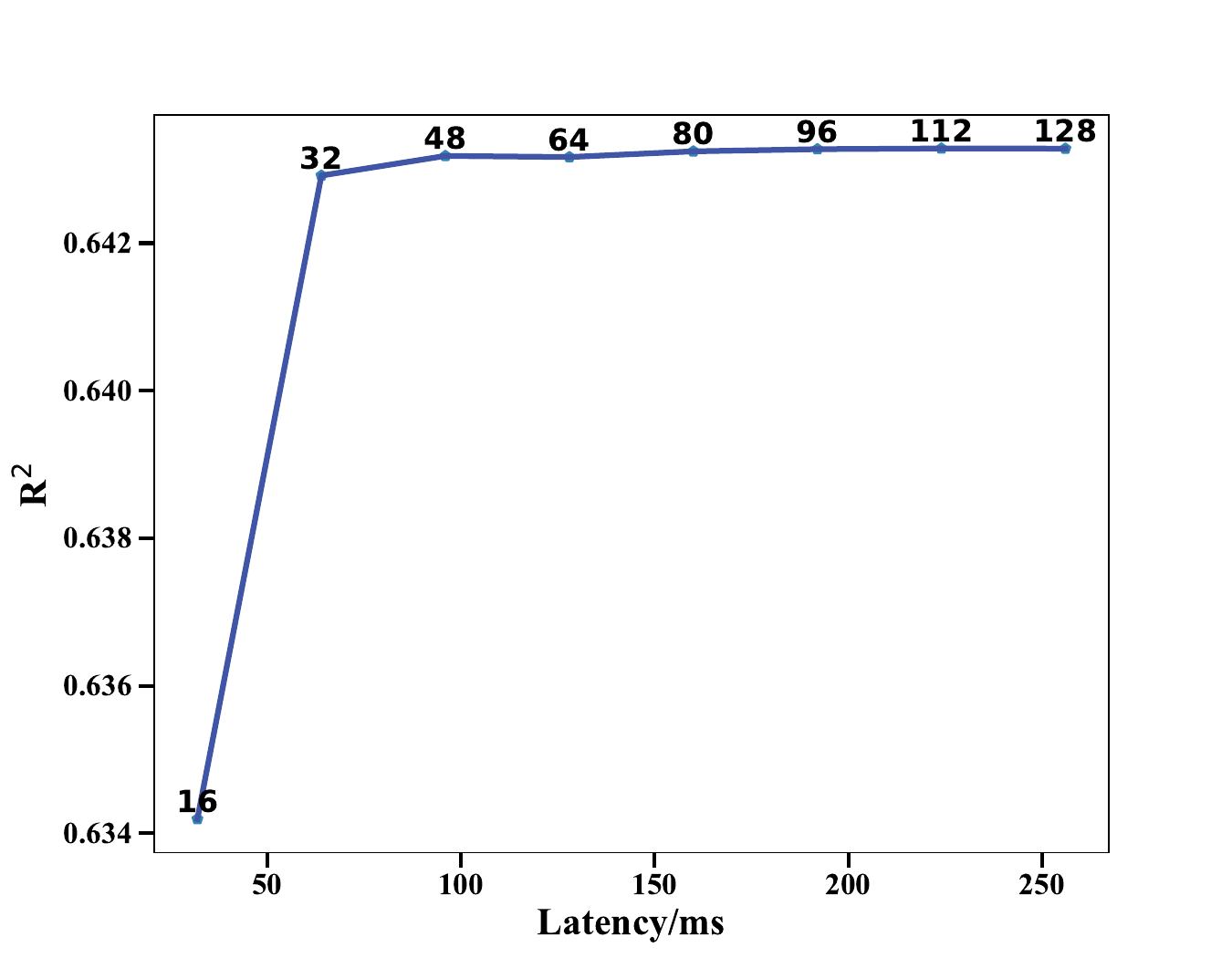}
    \caption{Latency versus $R^{2}$ for block Bid filtering using $2$-nd order Bessel filters with cut-off frequency of $0.05$.The number denotes different block sizes. Block size of $32$ gives an optimal tradeoff between accuracy and latency.}
    \label{fig:filter_latency}
\end{framed}
\end{figure}


First, we compare the performance of different types of filters; results for bidirectional filtering are shown here but similar conclusion holds for the Block bidirectional case as well. The comparison of these three filters with the original result is depicted in \Cref{fig:filter_compare}. Here, the order of the filter is fixed at $4$ and the cut-off frequency is $f_c=0.05$ based on the estimate from \Cref{fig:output_frequncy}. It is clear that the Bessel filter provides the maximum improvement of $R^2$ both in SNN\_3D and SNN\_Streaming model (statistical significance tested using paired t-test), which is about 8.2\% higher than SNN\_3D and 5.7\% higher than SNN\_Streaming without using filters. The constant delay property of Bessel filters is crucial in not distorting the waveform shape. To visualize this qualitatively, X and Y velocities for one reach from one file after filtering by the three filters is shown in \Cref{fig:filters_details}(reaches from other files are shown in the Supplementary material). It can be seen that the velocity trajectories after Chebyshev filtering is not as smooth as the other two with Bessel filters producing the smoothest and most natural trajectories. Thus, we choose Bessel filters to perform the rest of the experiments in this paper.

To find the combination of filter order and block size with the best trade-off in terms of accuracy and latency, we evaluated the performance for SNN\_3D and SNN\_Streaming using block Bid filtering of bessel filter with the cut-off frequency of $0.05$. The block size is varied between $16-80$, while the filter order is swept between $2-8$. The results for both types of SNN models shown in \Cref{fig:order_vs_block_size} indicate a large increase in $R^2$ when the block size increases to $32$ and marginal improvements from there on, making block size of $32$ a good choice. For block sizes of $32$, a filter order of $4$ is optimal (statistically significant difference between order 2 and order 4 tested at the $5\%$ level using paired t-test for both SNN\_3D and SNN\_Streaming), while for higher block sizes, filter order of $6$ gives better results. There are no results for block size of $16$ with filter order larger than $4$ because of insufficient samples to perform the filtering. In this case, filter order of $2$ gives the best accuracy (statistically significant difference between order 2 and order 4 exists using the paired t-test at the $5\%$ level for SNN\_Streaming and $10\%$ level for SNN\_3D). From these results, a block size of $32$ and a filter order of $4$ seems like the best choice. However, there is a direct tradeoff between filter order and latency of the output as discussed next; this may make block size of $16$ more appropriate for some applications\cite{hueber2024benchmarking}. 

Latency between input and output is important for real-time applications with closed-loop operation such as motor decoding. The total time, $T_{tot}$, taken to produce an output by a NN decoder is given by $T_{tot}=St+T_{comp}$ where $St$ is the stride to capture the new input data ($=T_{GT}=4$ ms in this work) and $T_{comp}$ is the time taken to process the computations in the neural network. Given the very fast and energy-efficient In-memory computing (IMC) approaches to implement NN models prevalent now\cite{abu_nvm,marvin_sram} and the small networks considered in this paper,  we can assume $T_{comp}<<St$ making the throughput almost entirely dependent on $St$, i.e. time taken to capture new neural input spikes. Note that the bin window, $T_W$ does not add any extra penalty on latency of output generation; however, after every change of target, the prediction will be inaccurate for a time related to $T_W$ to allow enough relevant input to fill up the bin window. However, output filtering may induce an extra penalty on the latency. Bid filters produced best results as seen in the \Cref{tab:results_summary}; however, they cannot be employed in real-time applications since they need to store the raw data in memory first and then apply forward filtering two times in opposite directions. The block Bid filter is chosen as a compromise where the filter window is used to determine the length or block of samples that are filtered at one time, and the predicted point is located at the center of the sample window. Thus, the latency introduced by the block Bid filter is theoretically equal to half the length of the filter window. \Cref{fig:filter_latency} compares the accuracy of 2nd order Bessel filters for various block sizes. This shows that the minimum latency achievable by a 2nd order Bessel filter with block size $16$ is $32$ ms for this dataset with $T_{GT}=4$ ms. Hence, we select two Bessel filter configurations for further simulations--\textbf{lower latency} (2nd order, block size $16$) and \textbf{higher accuracy} (4th order, block size $32$).

\begin{figure}[!t]
\begin{framed}
    \centering
    \includegraphics[width=0.65\textwidth]{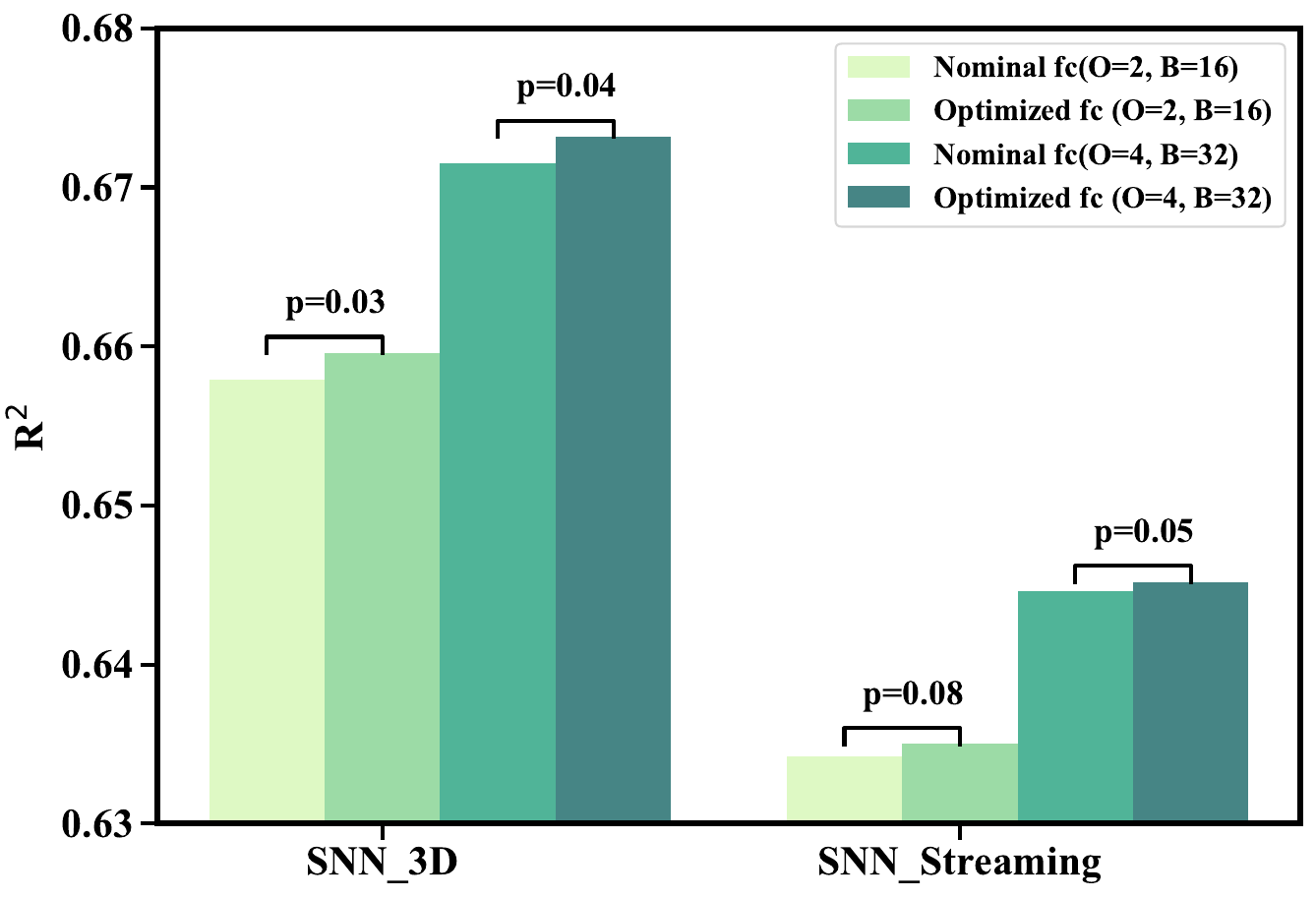}
    \caption{Cut-off frequency optimization for the two cases: one case of filter order of 2 and block size of 16, another case of filter order of 4 with block size of 32, respectively. Statistical
significance tested using paired t-test.}
    \label{fig:fc_optimization}
\end{framed}
\end{figure}

Lastly, we have so far chosen the cutoff frequency $f_c$ in an adhoc fashion based on \Cref{fig:output_frequncy}. However, $f_c$ also can be optimized for the above two configurations. SNN models are not directly involved in the training process; rather, the input data is processed by pre-trained SNNs and the model output is then passed to the Bessel filter. Two approaches were tried--one where the digital filter is treated as a single layer neural network and another where $f_c$ is directly optimized by using search algorithms, with the latter providing better results. Grid search is used first to determine a reasonably accurate range around $0.05$ for $f_c$ with the potential maximum of $R^{2}$. Then search step and range are halved, based on the preferred range found by grid search. This process is repeated independently for each of the six files until $R^2$ stops increasing. \Cref{fig:fc_optimization} shows the optimization results for both SNN\_Streaming and SNN\_3D for the two configurations. The results indicate a statistically significant difference between filtered and unfiltered data exists using the paired t-test at the 5\% level for SNN\_3D and 10\% level for SNN\_Streaming.


\subsection{Baseline result and Pareto plots}
\label{subsec:baseline_result}
\begin{figure}[!t]
\begin{framed}
    \centering
    \includegraphics[width=0.95\textwidth]{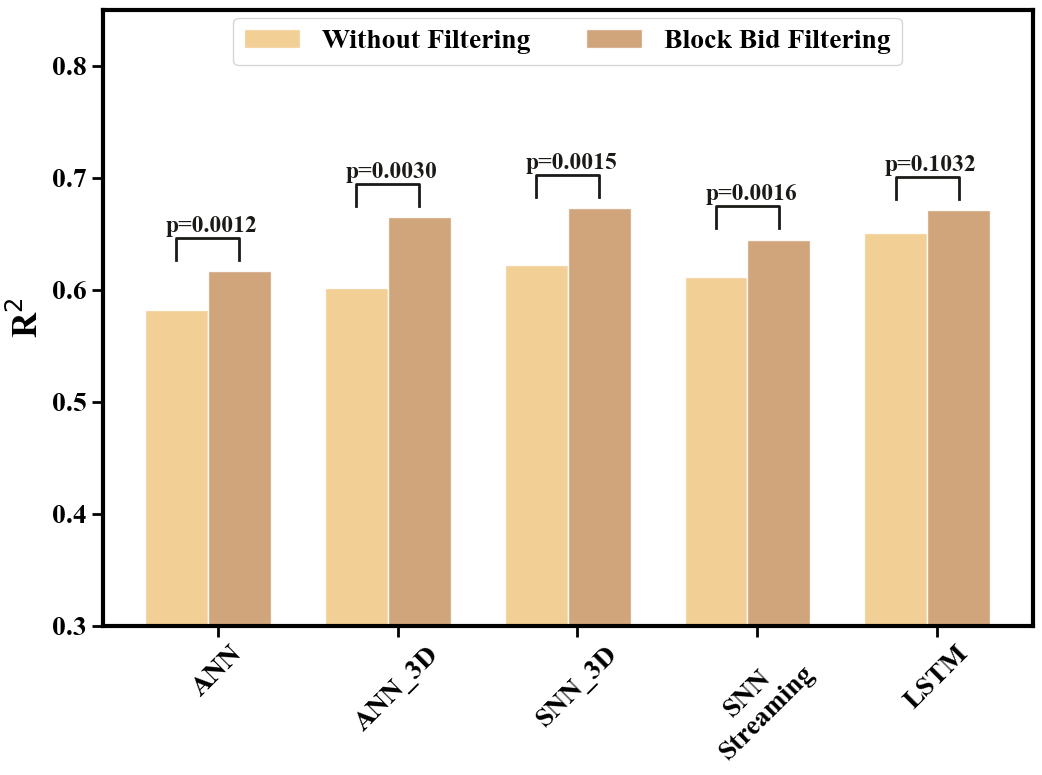}
    \caption{Comparison of baseline performance across various models, both using block bid filtering and without filtering. The statistically significant difference between filtered and unfiltered data exists using the paired t-test at the 5\% level for ANN and SNN models.}
    \label{fig:original_dataset_compare}
\end{framed}
\end{figure}

\begin{figure}[ht]
\begin{framed}
    \centering
    \begin{subfigure}[b]{0.45\textwidth}
        \centering
        \includegraphics[width=\textwidth]{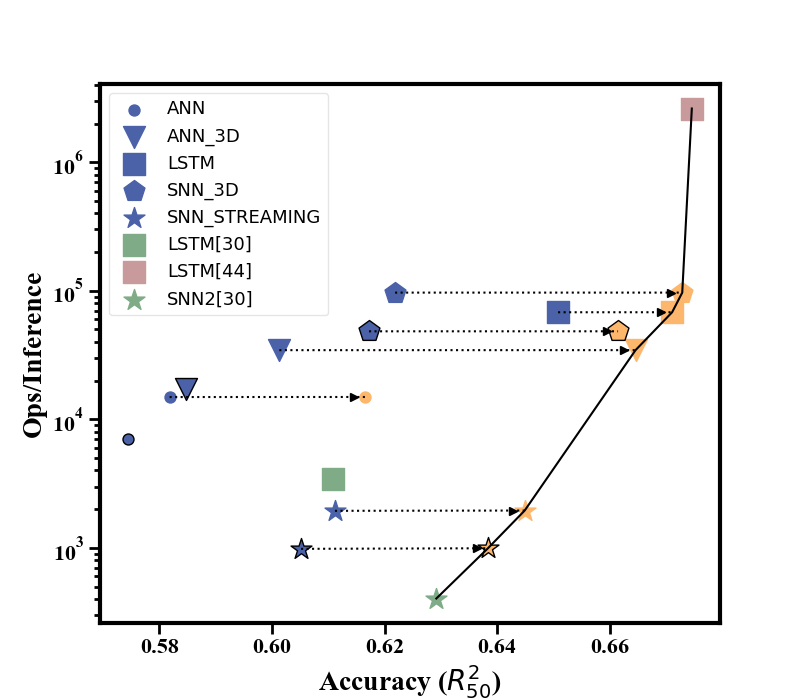}
        
    \end{subfigure}
    \hfill
    \begin{subfigure}[b]{0.45\textwidth}
        \centering
        \includegraphics[width=\textwidth]{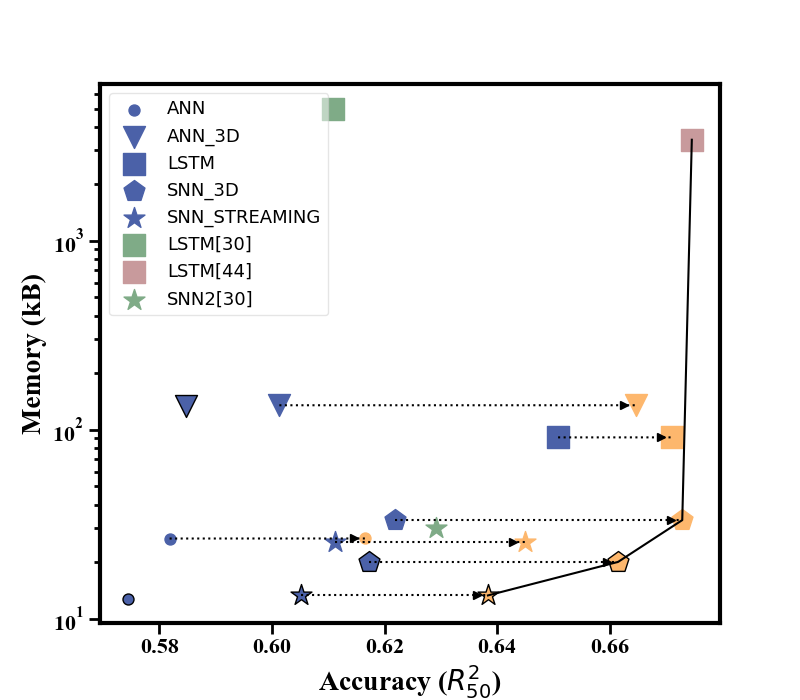}

    \end{subfigure}

    \caption{Baseline Pareto plots using $50\%$ training data showing trade-offs for different models: a) Compute cost vs. accuracy b) Memory footprint vs. accuracy. The following colour scheme is used: (1) blue markers are base models without filtering (correspond to results from prior work in \cite{Yik2023} for ANN, ANN\_3D and SNN\_3D)  (2) orange markers are models using block Bid filtering with 4th order and block size 32. (3) markers with dark border are tiny variants of base models.
}
    \label{fig:baseline_result}
\end{framed}
\end{figure}

Different ANN and SNN models (baseline and tiny as explained in Section \ref{sec:Methodology}) are trained and evaluated on the dataset; the results for the baseline variants are detailed in \Cref{tab:results_summary}. The \textbf{high accuracy} configuration of filter is used here, i.e. the filter order and cutoff frequency are $4$ and $0.05$ for bidirectional filtering, while in block bid filtering, the block size is selected as $32$. Results for the \textbf{low latency} configuration are shown in the Supplementary data. 

Computes and model size are obtained from the Neurobench code harness\cite{Yik2023} \cite{yik2023neurobench}. In NeuroBench, computes are broken down into the following three types: dense, effective Multiply-Accumulates (Effective MACs),  and effective Accumulate Synaptic operations (Effective ACs). Dense computes accounts for all zero and nonzero neuronal activations and synaptic connections. This is used to reflect the number of operations needed on hardwares that does not support sparse operations. Effective MACs and effective ACs only take into consideration of operations that are nonzero, i.e. any zero activation by ReLUs, no spike output by SNNs, or zero synaptic connections are ignored, reflecting the operations that would take place on hardwares that support sparsity. NeuroBench computes a model’s footprint by taking the following into consideration: quantization level of the weights, parameters (Weights and Biases), and buffers needed for preprocessing of input data (for a realistic inference comparison). Note for a model’s footprint, zero weights are included as well, as they are part of the connection sparsity metrics.

For a fair comparison across models, we have also added filtering to the output of other models such as ANN and LSTM. The general trend observable from \Cref{tab:results_summary} and \Cref{fig:original_dataset_compare} is that filtering improves the $R^2$ for all the NN models. Compared to the NN models, signal processing methods like SSKF have much lower computes but the $R^2$ is significantly lower due to its inability to track changes in data by varying KF gain\cite{kalman1960new,malik2010efficient}. On the other hand, UKF or rEFH have much higher computations due to matrix inversions\cite{hueber2024benchmarking}, but still do not attain similar $R^2$ as the NN methods. The baseline accuracy we obtained was 0.607 for SNN2 \cite{hueber2024benchmarking}, which can be improved to 0.6354 with block filtering. Compared to earlier LSTMs\cite{hueber2024benchmarking, glaser2020machine} and SNNs\cite{hueber2024benchmarking}, the filtered LSTM and SNN\_Streaming in this work achieves higher $R^2$ with less computes and memory. To visually compare the tradeoffs between accuracy and resource usage, we use pareto plots (\Cref{fig:baseline_result}).  The black line is the Pareto frontier, which indicates the best trade-off between the accuracy and operations/memory, the ideal place being the lower right corner of the plot. To keep the plots less cluttered, we only plot results with block Bid filtering since Bid filtering cannot be used in real-time implementations in any case. We also add the tiny variants in the plots. For the pareto plots shown in this section, the following colour scheme is used: 
\begin{itemize}
    \item blue markers are base models without filtering (correspond to results from prior work in \cite{Yik2023} for ANN, ANN\_3D and SNN\_3D) 
    \item orange markers are models using block Bid filtering with block size of 32 and order of 4.
    \item markers with dark border are tiny variants of base models
\end{itemize}

\begin{figure}[!t]
\begin{framed}
    \centering
    \begin{subfigure}[b]{0.45\textwidth}
        \centering
         \includegraphics[width=\textwidth]{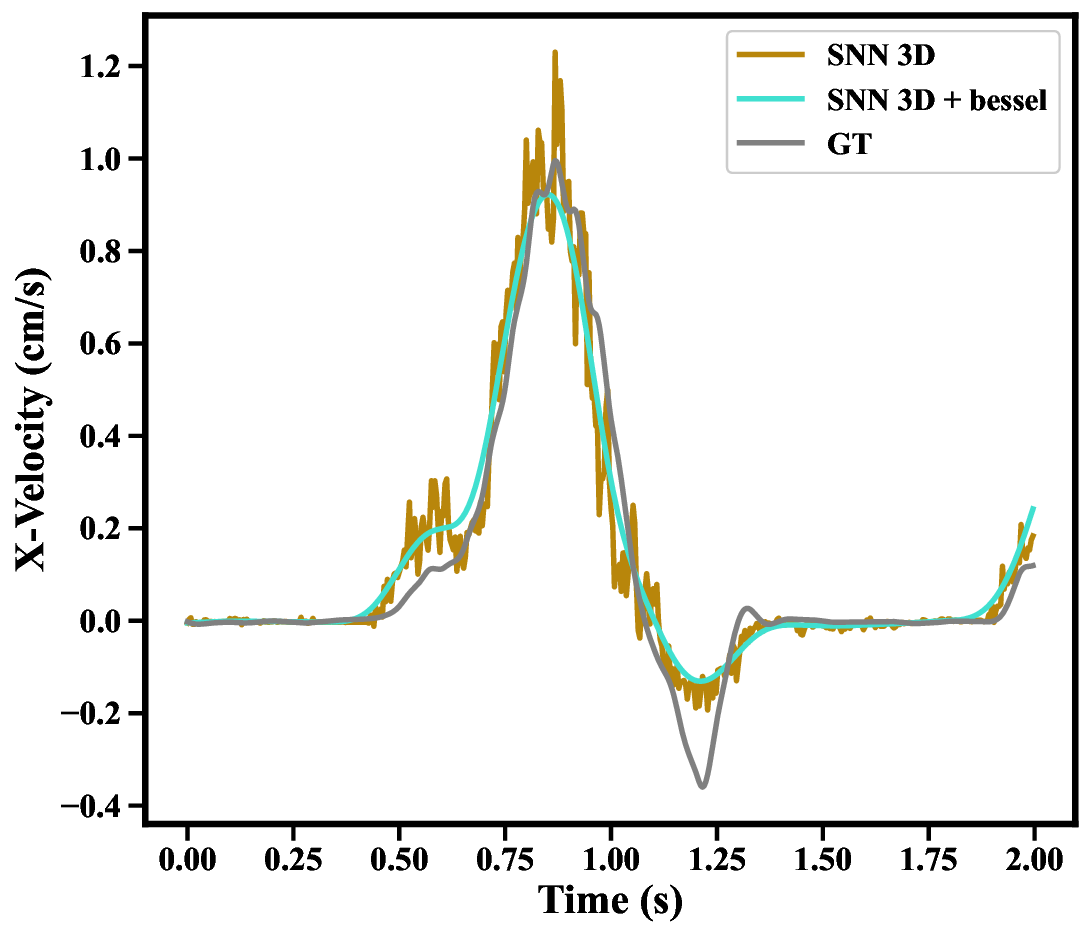}
        \caption{}
    \end{subfigure}
    \hfill
    \begin{subfigure}[b]{0.45\textwidth}
        \centering
         \includegraphics[width=\textwidth]{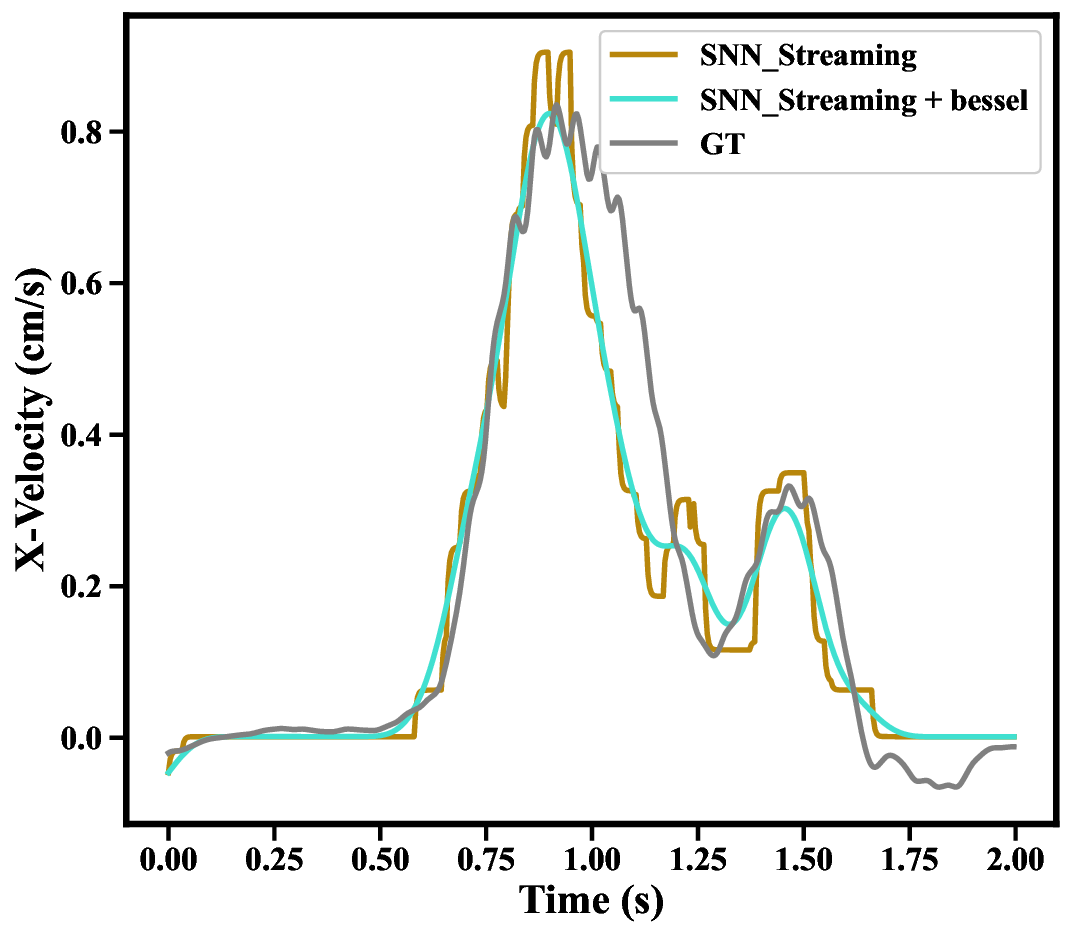}
        \caption{}
    \end{subfigure}
    \caption{Predicted trajectory of X-velocity with and without filter showing the smoothness introduced by the filter making it similar to natural motion: a) SNN\_3D. b) SNN\_Streaming}
    \label{fig:filter_vs_prediction}
\end{framed}
\end{figure}

First, we compare the performance using the $50\%$ data split as done in \cite{Yik2023} using pareto plots as shown in \Cref{fig:baseline_result}. In terms of the models that forms the pareto front of operations vs. accuracy (\Cref{fig:baseline_result}(a)), we observe that filtered SNNs dominate with \textit{SNN\_3D} occupying the higher part while \textit{SNN\_Streaming} occupying the lower part. These results can be taken as a gold standard for the neurobench suite\cite{Yik2023} at this time since they represent the highest reported accuracy so far. The two SNNs variants show a big difference in terms of operations required ($\approx 100$x) and accuracies ($\approx 4\%$). The block filtering increased the accuracy of \textit{SNN\_3D} and \textit{SNN\_Streaming} by 8\% and 5.7\% with only a 0.015\% and 0.512\% increase in computes, respectively. \Cref{fig:filter_vs_prediction} plots the actual trajectory of a ground truth reach waveform, a prediction each from \textit{SNN\_3D} and \textit{SNN\_Streaming}, and the corresponding filtered versions. It can be seen how the filtered waveform is smoother and resembles the more natural motion of the primate's finger. Also. outputs from \textit{SNN\_Streaming} are inherently smoother than \textit{SNN\_3D} due to its internal memory through membrane potentials which do not get reset like \textit{SNN\_3D}. In terms of memory usage (\Cref{fig:baseline_result}(b)), the pareto front is also dominated by filtered SNNs. Here, block filtering usage only increases model size by a mere 0.058\% and 0.077\% for \textit{SNN\_3D} and \textit{SNN\_Streaming}. The \textit{ANN\_3D} models have highest memory usage due to their input dimension being expanded by $m$ times to $m\times N_{ch}$--the weights in the first layer are dominant for memory footprint since $N_0>>N_1,N_2,N_3$.


Looking deeper at the effects of filtering, we see that \textit{SNN\_Streaming} with block Bid filtering achieves similar accuracy of $\approx 0.64$ as the \textit{LSTM} without filtering at $\approx 27\%$ memory and $\approx 25\%$ computations. This confirms our initial hypothesis that adding memory via filtering to SNN models can indeed make their performance similar to recurrent ANNs. Even the tiny variant of \textit{SNN\_3D} model achieves higher accuracy with slightly less operations and only $30\%$ memory usage compared to \textit{LSTM}. In summary, filtered SNNs are the best performing models and either \textit{SNN\_3D} or \textit{SNN\_Streaming} may be chosen depending on the desired tradeoff between accuracy and resource usage.

\subsection{80\% vs 50\% Training Split}
\label{subsec:increase_data}
\begin{figure}[!t]
\begin{framed}
    \centering
    \includegraphics[width=0.45\textwidth]{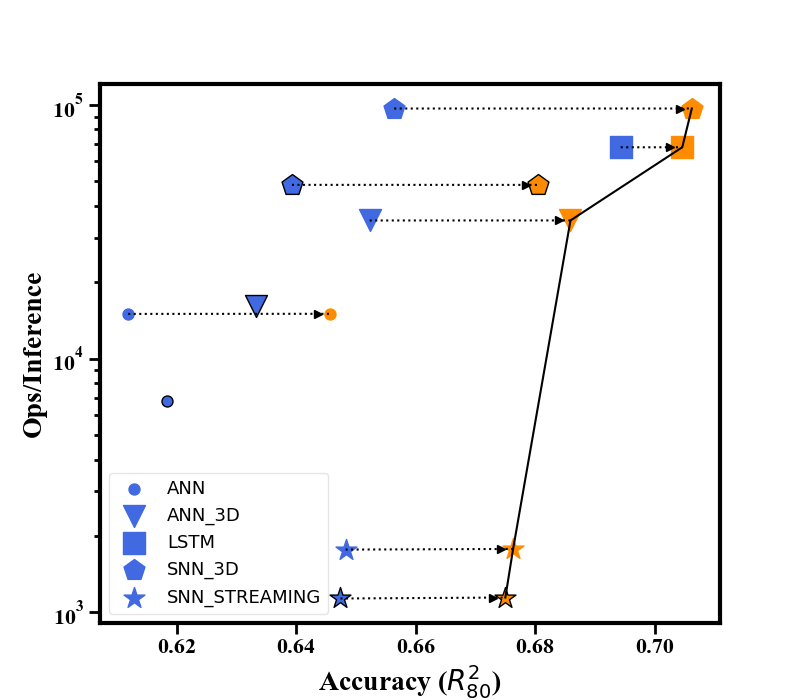}
    \hfill
    \includegraphics[width=0.45\textwidth]{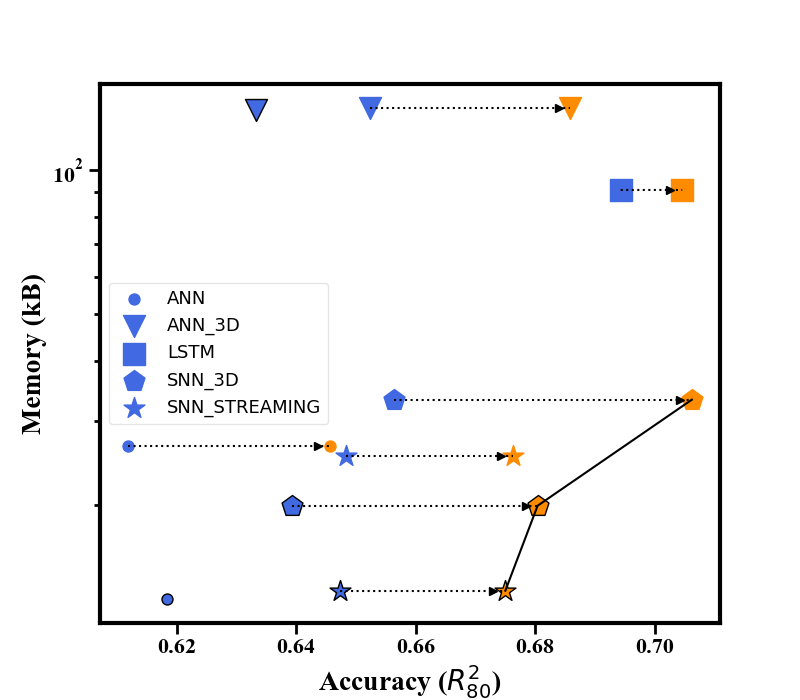}
  
    \caption{Pareto plots showing trade-offs for different models with increased training data to $80\%$ from the originally used $50\%$: a) Compute cost vs. accuracy b) Memory footprint vs. accuracy. The following colour scheme is used: (1) blue markers are base models without filtering  (2) orange markers are models using block Bid filtering with 4th order and block size 32. (3) markers with dark border are tiny variants of base models.}
    \label{fig:pareto_80_vs_50_split}
\end{framed}
\end{figure}

To assess the performance of models when the training data increases, we increase the baseline training data from 50\% to 80\% as done in \cite{Liao2022}. The results are listed in \Cref{tab:results_summary} and plotted in \Cref{fig:pareto_80_vs_50_split}. As expected, the $R^{2}$ of all models is generally higher by $0.03-0.04$ (written in parenthesis in \Cref{tab:results_summary}) compared to the 50\% baseline training data, which shows a high capacity for future improvement with more data. Similar to \Cref{fig:baseline_result},  both the pareto curves in \Cref{fig:pareto_80_vs_50_split} for computations and memory are dominated by SNNs with block Bid filtering. However, one filtered LSTM model and one filtered ANN\_3D model are now placed on the pareto frontier for computations. 

\section{Discussion}
\label{sec:Discussion}
This section discusses additional control experiments and gives an outlook for future improvements.

\subsection{Effect of Reach Removal}
As mentioned in \Cref{subsec:training_details}, some of the reaches in the dataset spanned a much longer duration (sometimes longer than 200 seconds) than the rest which mostly were less than 4 seconds. These reaches (longer than 8 seconds) were removed from training since the NHP was likely unattentive in these cases. However, they were not removed from the testing data and hence, we explored how much improvement in performance is obtained by better curating the test dataset. These results are presented in the \Cref{tab:ReachRemoval} and we can observe that the $R^{2}$ increases by $\approx 0.01$ with the baseline 50\% split--the improvement can be much more if other files from \cite{dataset} are selected. This underlines the effectiveness and necessity of careful data selection from the recordings in \cite{dataset} while training and testing models. 

\begin{table}[t]
\centering
\caption{Effect of Reach Removal--increase in $R^2$ over baseline for \textbf{high accuracy} filter configuration shown in parenthesis}
\centering
\footnotesize
\begin{tabular}{@{}lllllll}
\br

\centre{1}{\textbf{Models}} & & \centre{1}{\textbf{Split}} & & \centre{1}{\textbf{Filters}} & & \centre{1}{\textbf{R$^{2}$}} \\

\mr

\centre{1}{ANN} & & \centre{1}{50\%} && \centre{1}{No Filter} && \centre{1}{0.5921 $($+0.01$)$}\\
 
\mr

\centre{1}{ANN\_3D} & & \centre{1}{50\%} && \centre{1}{No Filter} && \centre{1}{0.6133 $($+0.01$)$}\\
 
\mr

\centre{1}{SNN\_3D} & & \centre{1}{50\%} && \centre{1}{No Filter} && \centre{1}{0.6286 $($+0.007$)$}\\
 
&&&\crule{4}\\

& &  && \centre{1}{Block Bid Filtering} && \centre{1}{\textbf{0.6784 $($+0.006$)$}}\\
 
&&&\crule{4}\\

& &  && \centre{1}{Bid Filtering} && \centre{1}{0.6755 $($+0.007$)$}\\
 
\mr

\centre{1}{SNN\_Streaming} & & \centre{1}{50\%} && \centre{1}{No Filter} && \centre{1}{0.6212 $($+0.01$)$}\\
 
&&&\crule{4}\\

& &  && \centre{1}{Block Bid Filtering} && \centre{1}{0.6532 $($+0.008$)$}\\
 
&&&\crule{4}\\

& &  && \centre{1}{Bid Filtering} && \centre{1}{\textbf{0.6542 $($+0.008$)$}}\\
 
\mr

\centre{1}{LSTM} & & \centre{1}{50\%} && \centre{1}{No Filter} && \centre{1}{0.6784 $($+0.03$)$}\\

\br
\end{tabular}
\label{tab:ReachRemoval}
\end{table}

\subsection{Generalization to other datasets}
\label{subsec:mc_maze}
To show the general applicability of our work, we have evaluated the proposed methods on another neural decoding dataset referred to as `MC Maze'\cite{churchland2022mc_maze,churchland2010cortical} that has been used to evaluate other SNN decoders \cite{taeckens2024spiking, krishna2023sparsity, zhang2024efficient}. Briefly, this dataset contains recordings from the motor and premotor cortex of a monkey as it performed delayed reaching tasks \cite{churchland2010cortical}. The reaches were either straight or curved to avoid virtual barriers. Neural data was recorded using two 96-electrode arrays implanted in the PMd and M1 regions. After an offline spike sorting, spike information of 107 neurons, hand position, and monkey gaze position in 1 ms bins are provided. Reaching tasks last up to 600 ms.

\begin{figure}[!t]
\begin{framed}
    \centering
    \includegraphics[width=0.95\textwidth]{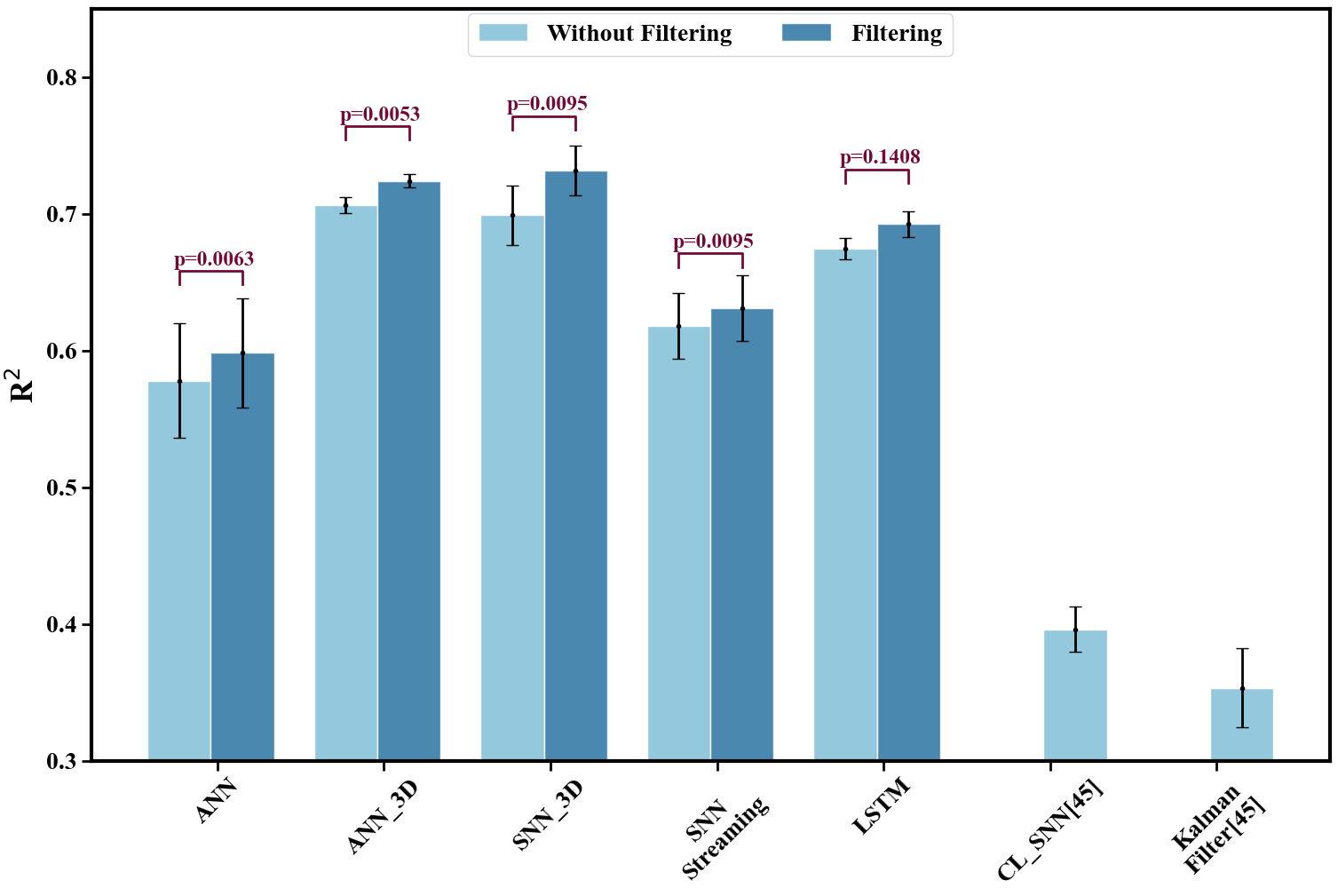}
    \caption{Comparison of  results for ANN, ANN\_3D, SNN\_3D, SNN\_Streaming and LSTM with CL\_SNN and Kalman filter on the MC Maze dataset. Except the LSTM, all other models show statistically significant improvement due to filtering.}
    \label{fig:mc_maze}
\end{framed}
\end{figure}

All the models described in the earlier sections were trained on this task. Decoders were trained on spiking data
from 130 ms prior to movement onset to 370 ms after
movement onset\cite{churchland2010cortical}. A block bidirectional filtering with a block size of 32 is implemented. \Cref{fig:mc_maze} displays the results of a comparison with the offline decoding results from \cite{taeckens2024spiking} using a Kalman Filter and a continuous learning SNN (CL\_SNN). To align with the results reported in \cite{taeckens2024spiking}, training data is set as 60\% while others are halved and set to testing/validation. It can be seen that all the decoding methods described here performed better than the earlier work\cite{taeckens2024spiking} (note that the X and Y velocity regressions are not shown separately as in \cite{taeckens2024spiking}, the final results are represented as the average of the X and Y directions (i.e. $R^2 = \frac{R_X^2+R_Y^2}{2}$), with the calculation details provided in the \Cref{subsec:metrics}). Moreover, filtering improves decoding accuracy in a statistically significant way (paired t-test) for all the models except LSTMs. SNN\_3D again achieves the best performance after filtering while SNN\_Streaming has the best tradeoff.

\subsection{New training methods}
\label{subsec:ccbr}
The results in the earlier works were obtained by training the models directly for regression using backpropagation. However, regression is generally considered a more challenging task compared to classification. Hence, there is potential for exploring other training techniques such as Cascade Classification Based Regresion (CCBR), that cast the regression problem into a framework for classification \cite{shaeri2022challengesccbr}. We describe the method briefly here, with details in \cite{shaeri2022challengesccbr}. The output space is divided into zones (different classes) and the first classifier predicts the sample should fall in which zone. This produces a regression result corresponding to the centroid of the zone. Following this, the next set of classifiers predicts the remaining error in regression. We did some preliminary investigation and tested the CCBR method in the same way as the previous models, and the result is shown in \Cref{tab:results_ccbr}, where the classifier is chosen as ANNs/SNNs instead of an SVM as in \cite{shaeri2022challengesccbr}. In most cases, the accuracy of CCBR is lower than that of the original model. Nevertheless, we still see the same two trends: (1) Block bid filtering can significantly improve accuracy. (2) SNN\_3D with filtering achieves best accuracy while SNN\_Streaming achieves best tradeoff between accuracy and resource usage.

\begin{table}[t]
\centering
\caption{The performance of CCBR training.}
\scriptsize
\begin{tabular}{@{}lllllllll}
\br
&&&&\centre{1}{\textbf{Activation}} &&&& \centre{1}{\textbf{Model Size}} \\

\centre{1}{\textbf{Models}} & \centre{1}{\textbf{Split}} & \centre{1}{\textbf{Filters}} & \centre{1}{\textbf{R$^{2}$}} & \centre{1}{\textbf{Sparsity}} &  & \centre{1}{\textbf{Computes}} &  & \centre{1}{\textbf{$($ kB $)$}} \\

&&&&&\crule{3}\\

&&&&&\centre{1}{\textbf{MACs}} &\centre{1}{\textbf{ACs}} &\centre{1}{\textbf{Memory}}\\

&&&&&&& \centre{1}{\textbf{Access}}\\

\mr

\centre{1}{ANN\cite{shaeri2022challengesccbr}} & \centre{1}{{50$\%$}} & \centre{1}{{No Filter}} & \centre{1}{{0.5020}} & \centre{1}{{0.7514}} & \centre{1}{{11424}} & \centre{1}{{0}} & \centre{1}{{9310.86}} & \centre{1}{{95.71}}\\

&  & \centre{1}{{Block Bid Filtering}} & \centre{1}{{0.5716}} & \centre{1}{{0.7514}} & \centre{1}{{11429}} & \centre{1}{{0}} & \centre{1}{{9315.86}} & \centre{1}{{95.72}}\\

\mr

\centre{1}{{ANN\_3D\cite{shaeri2022challengesccbr}}} & \centre{1}{{50$\%$}} & \centre{1}{{No Filter}} & \centre{1}{{0.5349}} & \centre{1}{{0.7348}} & \centre{1}{{55776}} & \centre{1}{{0}} & \centre{1}{{19478.228}} & \centre{1}{{442.22}}\\

&  & \centre{1}{{Block Bid Filtering}} & \centre{1}{{0.5995}} & \centre{1}{{0.7348}} & \centre{1}{{55781}} & \centre{1}{{0}} & \centre{1}{{19483.228}} & \centre{1}{{442.23}}\\

\mr

\centre{1}{{SNN\_Streaming\cite{shaeri2022challengesccbr}}} & \centre{1}{{50$\%$}} & \centre{1}{{No Filter}} & \centre{1}{{0.4717}} & \centre{1}{{0.7453}} & \centre{1}{{0}} & \centre{1}{{1942}} & \centre{1}{{2328.3}} & \centre{1}{{91}}\\

&  & \centre{1}{{Block Bid Filtering}} & \centre{1}{{0.5094}} & \centre{1}{{0.7453}} & \centre{1}{{0}} & \centre{1}{{1947}} & \centre{1}{{2333.3}} & \centre{1}{{91}}\\

\mr

\centre{1}{{SNN\_3D\cite{shaeri2022challengesccbr}}} & \centre{1}{{50$\%$}} & \centre{1}{{No Filter}} & \centre{1}{{0.5581}} & \centre{1}{{0}} & \centre{1}{{51744}} & \centre{1}{{0}} & \centre{1}{{65339.4}} & \centre{1}{{116.26}}\\

&  & \centre{1}{{Block Bid Filtering}} & \centre{1}{{0.6044}} & \centre{1}{{0}} & \centre{1}{{51749}} & \centre{1}{{0}} & \centre{1}{{65341.4}} & \centre{1}{{116.27}}\\

\br
\end{tabular}
\label{tab:results_ccbr}
\end{table}

\subsection{Future Directions}

The main reason for low energy consumption in SNN is due to the benefits of sparse activations. However, our experiment shows the sparsity may harm accuracy. We proposed two types of SNN models in this paper--one is SNN\_3D, which has no sparsity due to the layer normalization, and another one is SNN\_Streaming, which has a relatively higher sparsity. Interestingly, the low power characteristic of SNN is not reflected in the first SNN model, whereas it has relatively higher accuracy. This points to the need for future research into data normalization techniques which can still retain sparsity of activations. Another reason for the high accuracy of SNN\_3D was its reset of membrane potential after every $T_W$. This implies the membrane potential during training and testing start at exactly the same value for any sequence of inputs making it easier for the network to recognize similar patterns of input. For SNN\_Streaming, since there is no regular reset mechanism, the membrane voltages during training and testing may be quite different which may hurt accuracy. Mitigating this issue with initial condition of streaming SNNs will be a part of future work.

We see different models along the pareto curve having different strengths. For example, models with block Bid filtering have high accuracy but high latency. Using multiple models to produce a combined output may be a useful strategy. For example, switching from a model with block Bid filter to one without a filter right after a change of target/context will help in balancing latency and accuracy.

The results presented in this work were based on offline decoding while in real life, experiments are performed in a closed-loop mode with visual feedback being commonly used\cite{shoeb-closedloop,cunningham2011closed}. A software tool, online prosthetic simulator (OPS)\cite{cunningham2011closed} has been developed to emulate this closed-loop operation in real experiments and we will use this as our next step of exploration. Furthermore, the characteristics of the acquired signal also change over time due to scar tissue formation on electrodes or micro-motion of electrodes\cite{shoeb-tnsre} We will also investigate the possibility of using continuous learning \cite{taeckens2024spiking} to address the issue of data drift.

Finally, all the weights used in this work used float 32 as the default precision. However, there is a significant amount of work to quantize the models for more efficient inference. Applying these approaches of quantization aware training should allow us to reduce the model footprint significantly in the future.

\section{Conclusion}
Scaling iBMI systems to tens of thousands of channels in the future as well as removing the connecting wires would require significant compression of data on the device to reduce wireless datarates. Integrating the signal processing chain up to the neural decoder offers interesting opportunities to maximize compression. In this context, this work explores combining SNNs with traditional signal filtering techniques to improve their accuracy vs cost trade-offs where the cost is measured in terms of memory footprint and number of operations. Adding Bessel filtering improves the performance of both types of SNN models and block Bidirectional filtering generating the state-of-the-art results. Two filter variants for \textbf{high accuracy} and \textbf{low latency} are shown. In general, filtered \textit{SNN\_3D} and filtered \textit{SNN\_Streaming} models occupy the high and low ends of the pareto curves (for accuracy vs. memory/operations) respectively. Filtering the output of both ANN and SNN models with the high accuracy variant of Bessel filters exhibited statistically significant improvement in accuracy.

\section*{Acknowledgment}
We acknowledge useful discussions with the Motor decoding group in Neurobench. The work done in this paper was partially supported by a grant from the Research Grants Council of the Hong Kong Special Administrative Region, China (Project No. CityU 11200922).

\section{Supplementary Data}
\label{sec:supplementary_data}

\clearpage

\appendix
\counterwithin{figure}{section}
\renewcommand{\thefigure}{A.\arabic{figure}}

\subsection{Pareto plots for \textbf{low-latency} filter configuration with order 2 and block size 16}

\begin{figure}[ht]
\begin{framed}
    \centering
    \begin{subfigure}[b]{0.45\textwidth}
        \centering
        \includegraphics[width=\textwidth]{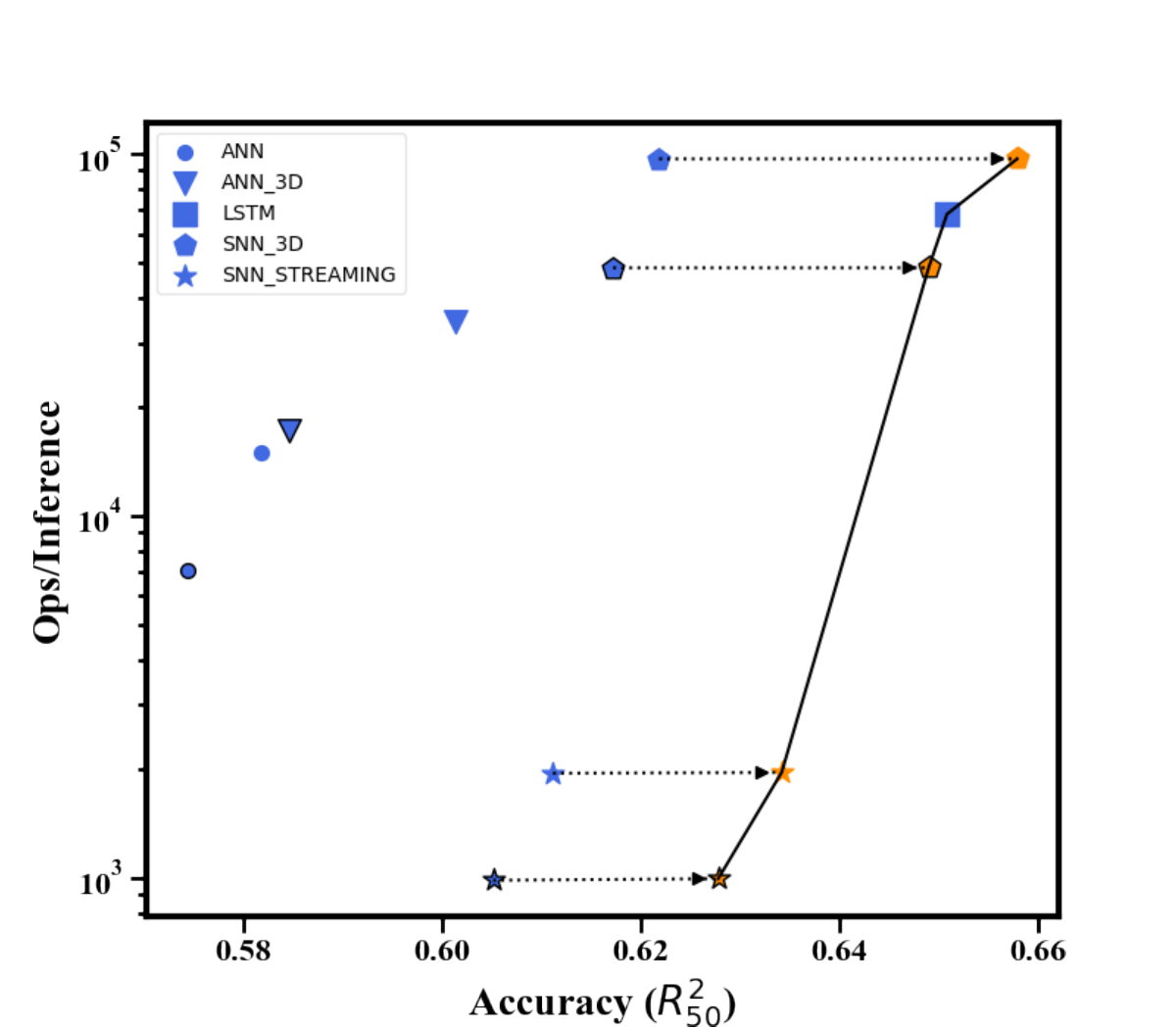}
    \end{subfigure}
    \hfill
    \begin{subfigure}[b]{0.45\textwidth}
        \centering
        \includegraphics[width=\textwidth]{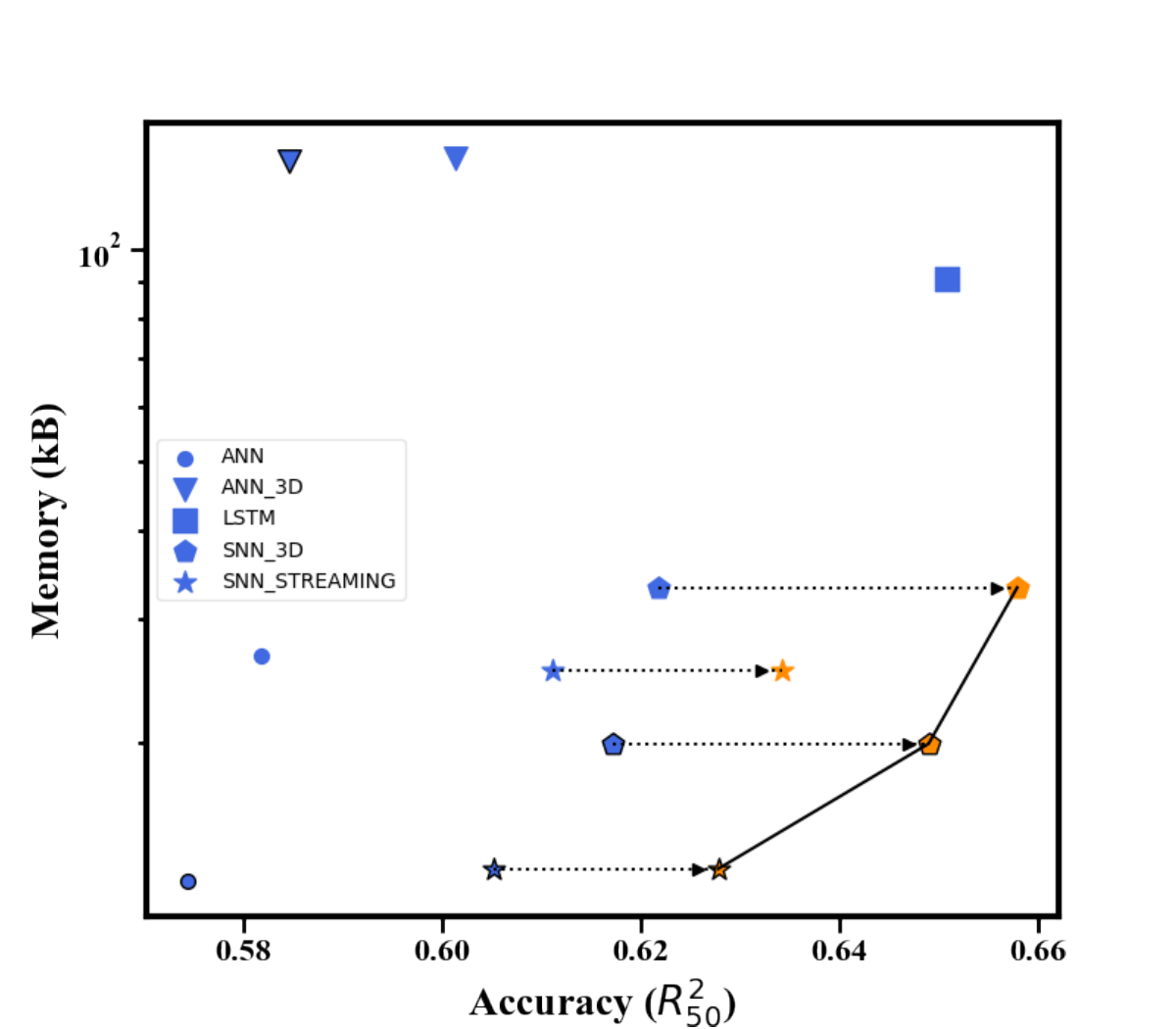}
    \end{subfigure}

    \caption{Pareto plots with \textbf{50\%} data split for \textbf{low$-$latency} Bessel filter configuration with block size of 16 and order of 2. (a) Compute cost vs. accuracy (b) Memory footprint vs. accuracy. SNNs with filtering dominate both pareto plots with one LSTM model appearing in the computation pareto. The following colour scheme is used: (1) blue markers are base models without filtering  (2) orange markers are models using block Bid filtering with 2nd order and block size 16. (3) markers with dark border are tiny variants of base models.}
    \label{fig:baseline_result_order2_block16_split50}
\end{framed}
\end{figure}

\begin{figure}[ht]
\begin{framed}
    \centering
    \begin{subfigure}[b]{0.45\textwidth}
        \centering
        \includegraphics[width=\textwidth]{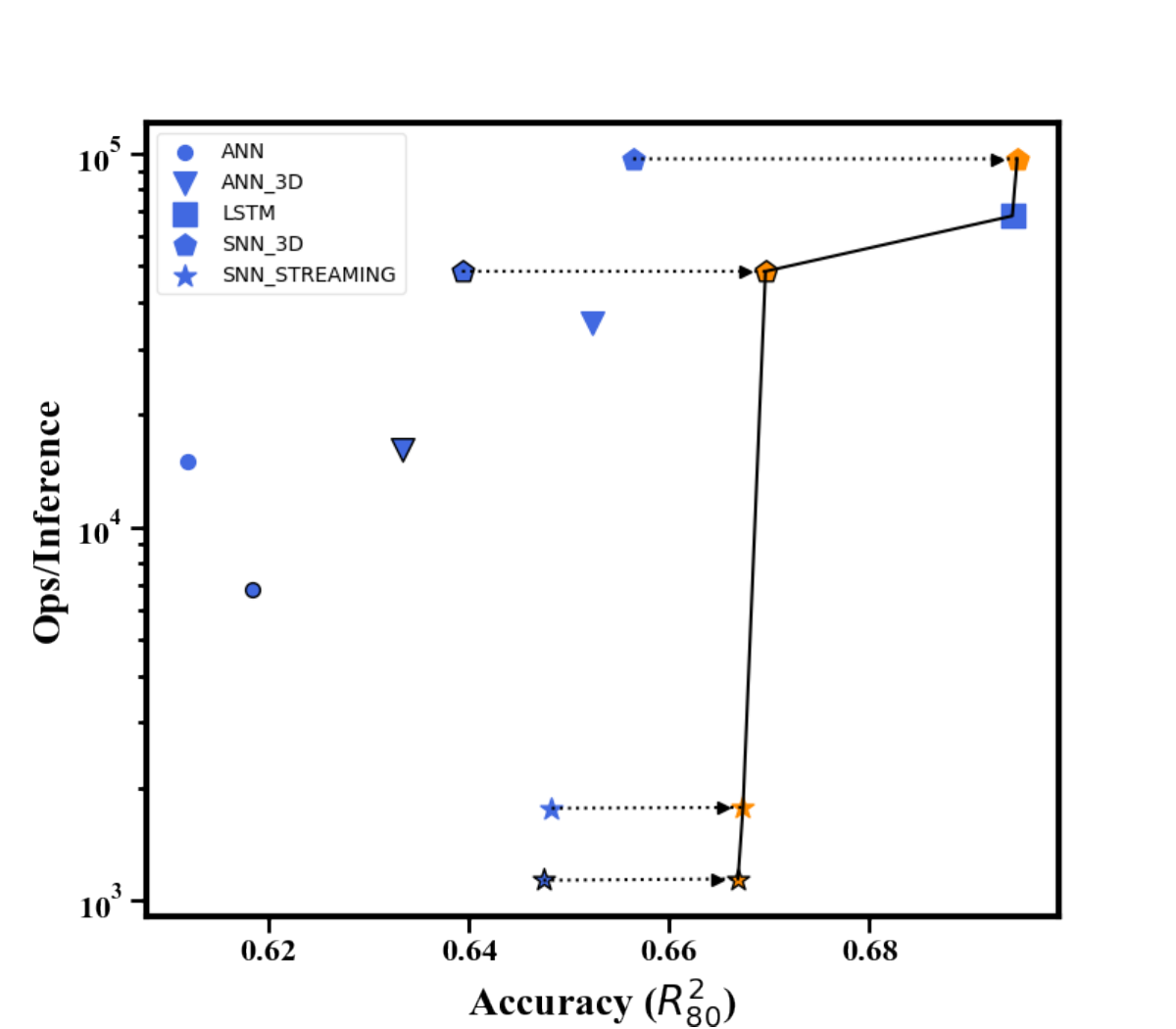}
    \end{subfigure}
    \hfill
    \begin{subfigure}[b]{0.45\textwidth}
        \centering
        \includegraphics[width=\textwidth]{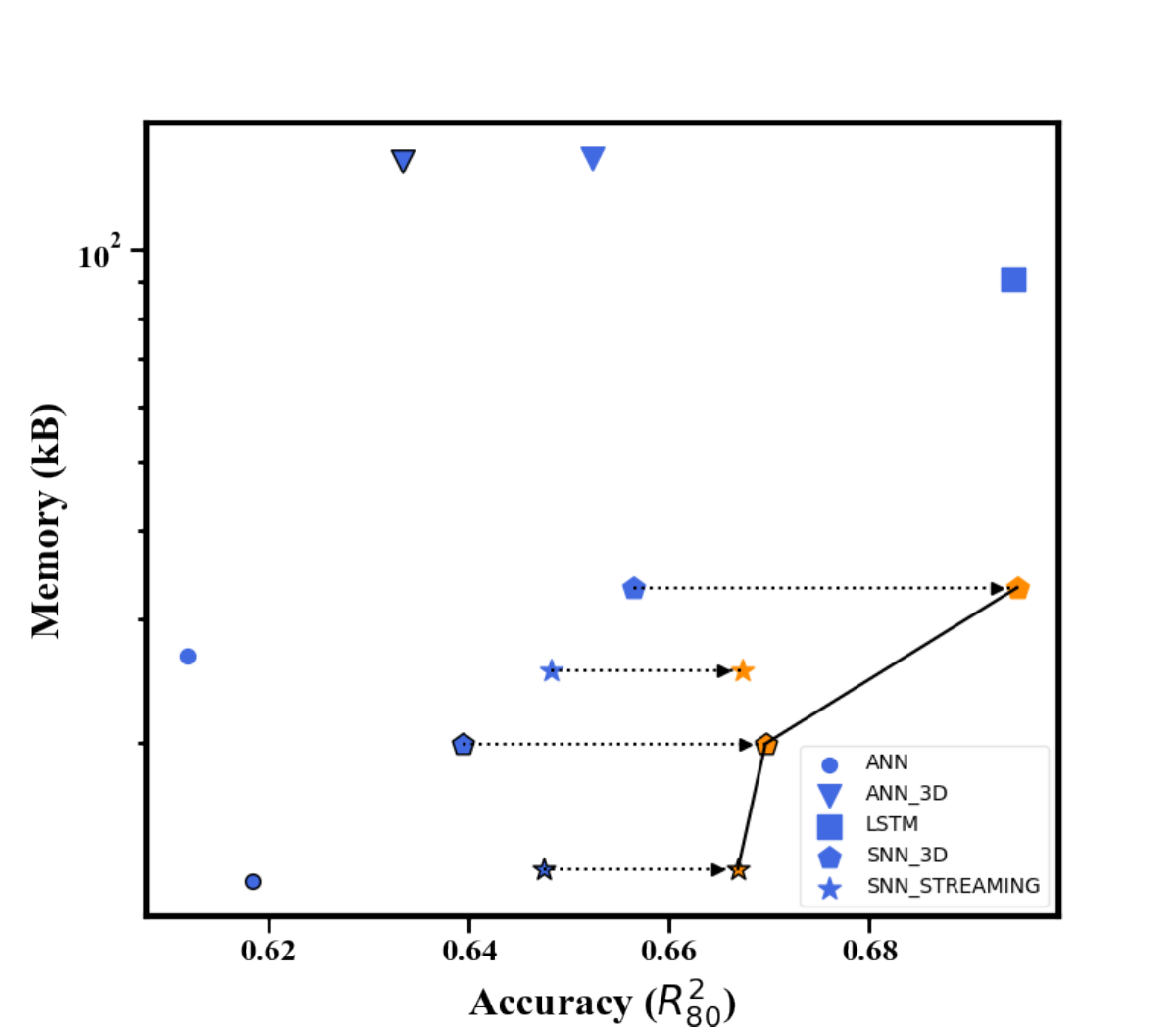}
    \end{subfigure}

    \caption{Pareto plots with \textbf{80\%} data split for \textbf{low$-$latency} Bessel filter configuration with block size of 16 and order of 2. (a) Compute cost vs. accuracy (b) Memory footprint vs. accuracy. SNNs with filtering dominate both pareto plots with one LSTM model appearing in the computation pareto. The following colour scheme is used: (1) blue markers are base models without filtering  (2) orange markers are models using block Bid filtering with 2nd order and block size 16. (3) markers with dark border are tiny variants of base models.}
    \label{fig:baseline_result_order2_block16_spilt80}
\end{framed}
\end{figure}

\clearpage

\subsection{Effect of 3 filters on 5 sample reaches from 5 different files}

\begin{figure}[ht]
\begin{framed}
    \centering
    \includegraphics[width=0.75\textwidth]{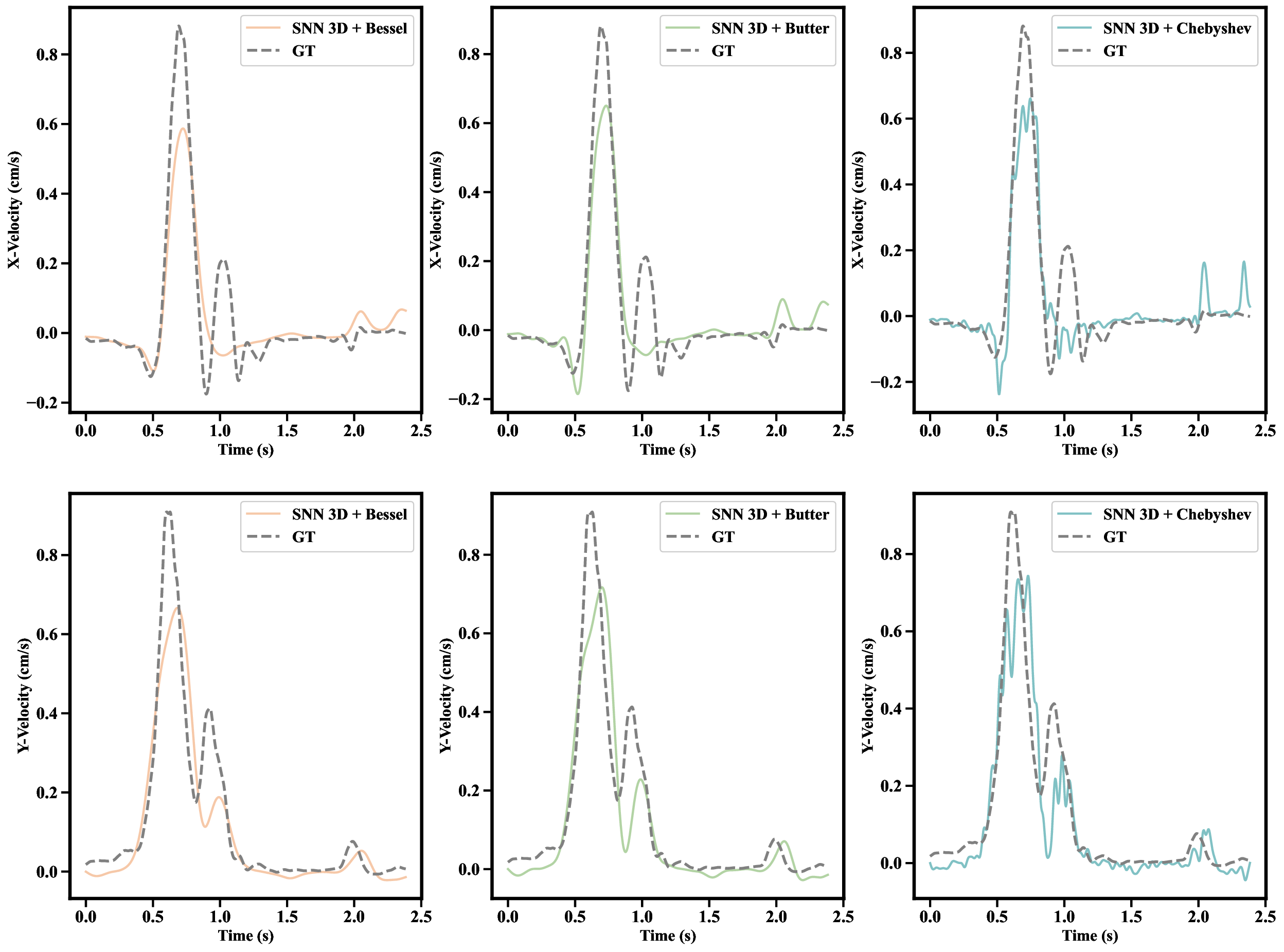}
    \caption{Comparison of three types of filters in terms of velocity and position based on one reach in the file $"indy\_20160630\_01"$. The three columns indicate SNN with Bessel, Butter, and Chebyshev filter respectively. The first row shows the X velocity, the second row shows the Y velocity. 4-th order filter with block size of 32 and cut-off frequency of 0.05 is used.}
    \label{fig:filters_details_indy_20160630_01}
\end{framed}
\end{figure}

\begin{figure}[ht]
\begin{framed}
    \centering
    \includegraphics[width=0.75\textwidth]{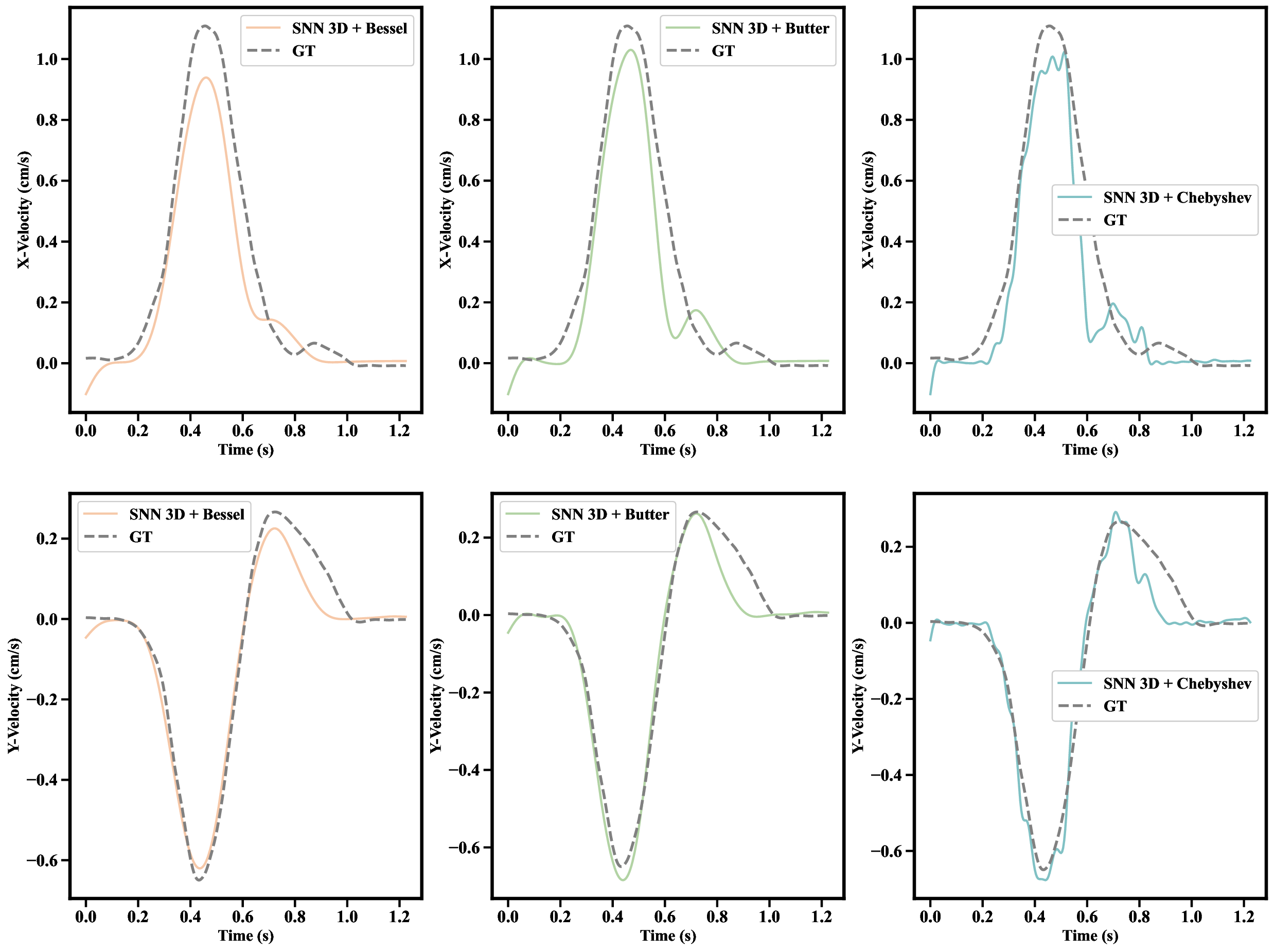}
    \caption{Comparison of three types of filters in terms of velocity and position based on one reach in the file $"indy\_20170131\_02"$. The three columns indicate SNN with Bessel, Butter, and Chebyshev filter respectively. The first row shows the X velocity, the second row shows the Y velocity. 4-th order filter with block size of 32 and cut-off frequency of 0.05 is used.}
    \label{fig:filters_details_indy_20170131_02}
\end{framed}
\end{figure}

\begin{figure}[ht]
\begin{framed}
    \centering
    \includegraphics[width=0.75\textwidth]{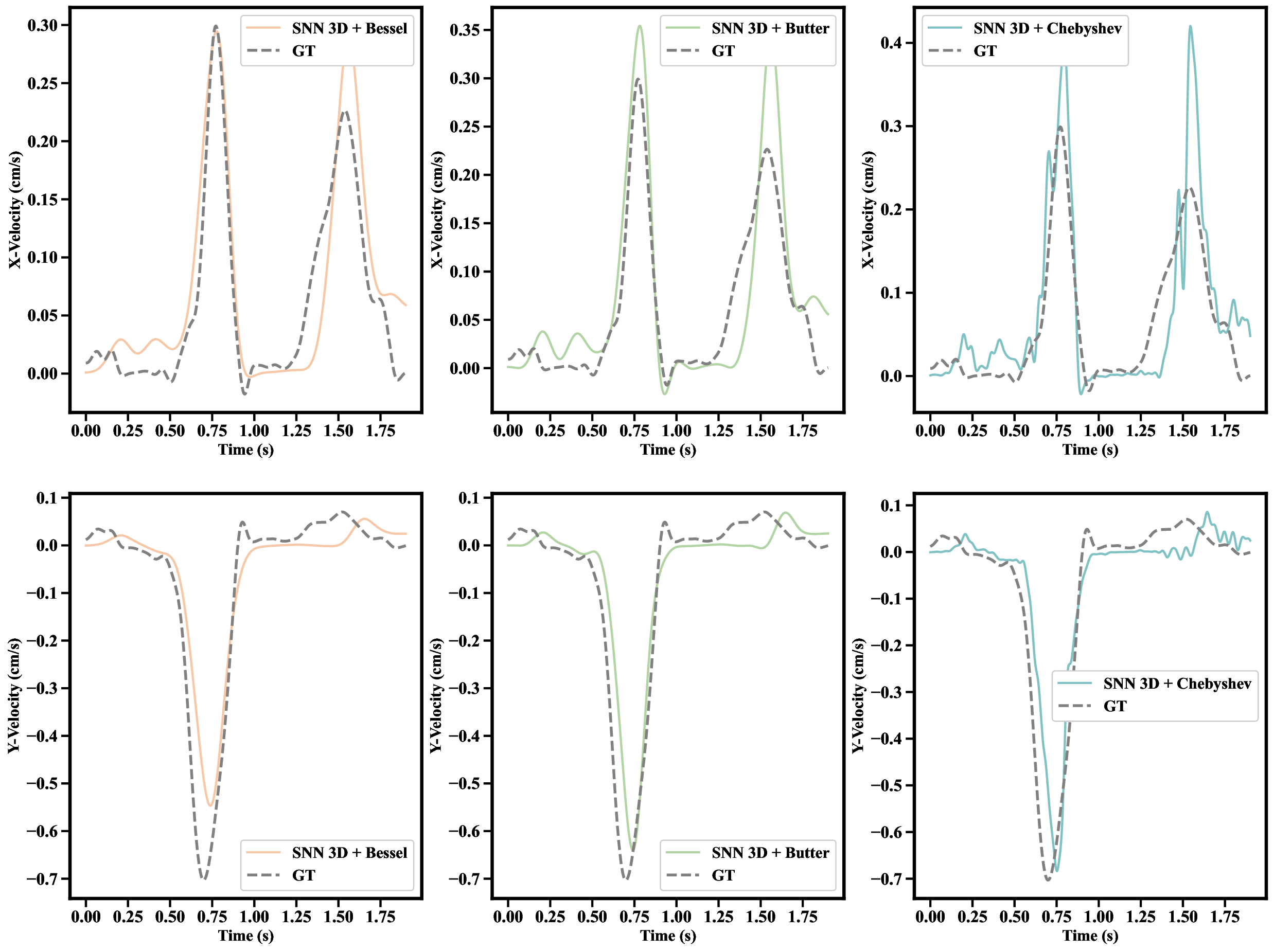}
    \caption{Comparison of three types of filters in terms of velocity and position based on one reach in the file $"loco\_20170210\_03"$. The three columns indicate SNN with Bessel, Butter, and Chebyshev filter respectively. The first row shows the X velocity, the second row shows the Y velocity. 4-th order filter with block size of 32 and cut-off frequency of 0.05 is used.}
    \label{fig:filters_details_loco_20170210_03}
\end{framed}
\end{figure}

\begin{figure}[ht]
\begin{framed}
    \centering
    \includegraphics[width=0.75\textwidth]{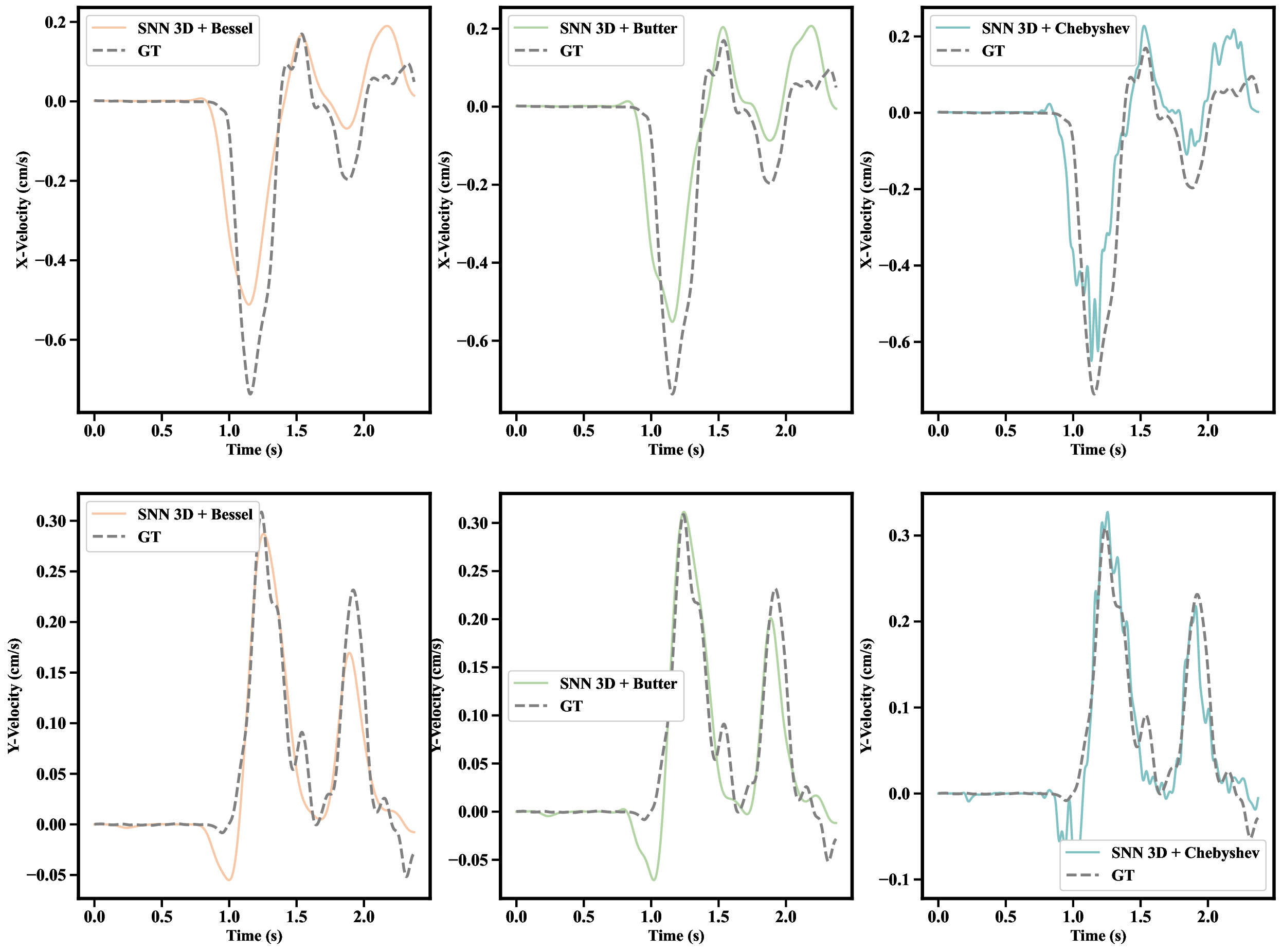}
    \caption{Comparison of three types of filters in terms of velocity and position based on one reach in the file $"loco\_20170215\_02"$. The three columns indicate SNN with Bessel, Butter, and Chebyshev filter respectively. The first row shows the X velocity, the second row shows the Y velocity. 4-th order filter with block size of 32 and cut-off frequency of 0.05 is used.}
    \label{fig:filters_details_loco_20170215_02}
\end{framed}
\end{figure}

\begin{figure}[ht]
\begin{framed}
    \centering
    \includegraphics[width=0.75\textwidth]{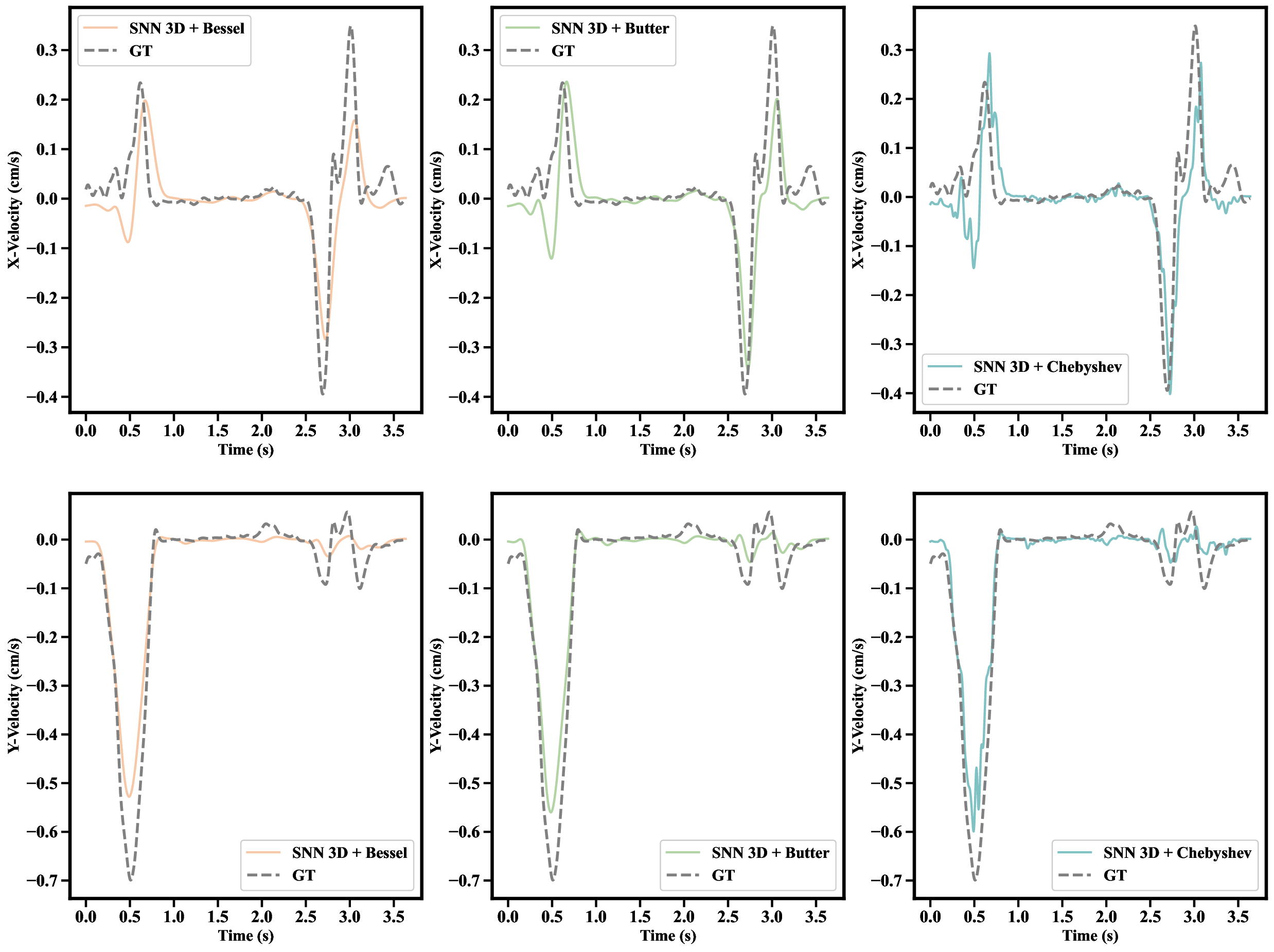}
    \caption{Comparison of three types of filters in terms of velocity and position based on one reach in the file $"loco\_20170301\_05"$. The three columns indicate SNN with Bessel, Butter, and Chebyshev filter respectively. The first row shows the X velocity, the second row shows the Y velocity. 4-th order filter with block size of 32 and cut-off frequency of 0.05 is used.}
    \label{fig:filters_details_loco_20170301_05}
\end{framed}
\end{figure}

\clearpage

\section*{References}
\bibliography{refs.bib}
\bibliographystyle{IEEEtran}

\end{document}